\documentclass[10pt,journal,compsoc]{IEEEtran}
\ifCLASSOPTIONcompsoc
\else
\fi
\ifCLASSINFOpdf
\else
\fi
\usepackage[cmex10]{amsmath}
\usepackage{algorithmic}
\usepackage{algorithm}

\usepackage{array}
\newcommand{\PreserveBackslash}[1]
{\let\temp=\\#1\let\\=\temp}
\newcolumntype{C}[1]{>{\PreserveBackslash\centering}p{#1}}
\newcolumntype{L}[1]{>{\PreserveBackslash\raggedright}p{#1}}

\usepackage{graphicx}
\usepackage{caption}
\usepackage{enumitem}

\usepackage{epstopdf}
\def\eg{{e.g.,\ }}
\def\ie{{i.e.,\ }}
\def\etal{{et al.\ }}
\def\n{{\mathbf{n}}}
\def\u{{\mathbf{u}}}
\def\v{{\mathbf{v}}}
\def\L{{\mathbf{L}}}
\def\R{{\mathbf{R}}}
\def\S{{\mathbf{S}}}

\def\I{{\mathbf{I}}}

\begin{document}
%
\title{Consistency-aware Shading Orders Selective Fusion  for Intrinsic Image Decomposition}
%
%
%
%

\author{Yuanliu~Liu, Ang~Li,
        ~Zejian~Yuan$^*$,~\IEEEmembership{Member,~IEEE},
        ~Badong~Chen,~\IEEEmembership{Senior~Member,~IEEE},
        and~Nanning~Zheng,~\IEEEmembership{Fellow,~IEEE}
\IEEEcompsocitemizethanks{\IEEEcompsocthanksitem Y. Liu, L. Ang, Z. Yuan, B. Chen and N. Zheng are with the Institute of Artificial Intelligence and Robotics, Xi'an Jiaotong University, Xi'an,
Shaanxi, 710049 China. Z. Yuan is the corresponding author.\protect\\
E-mail: liuyuanliu88@gmail.com, bennie.522@stu.xjtu.edu.cn\protect\\
\{yuan.ze.jian,chenbd,nnzheng\}@mail.xjtu.edu.cn}}

\IEEEtitleabstractindextext{%
\begin{abstract}
We address the problem of decomposing a single image into reflectance and shading. The difficulty comes from the fact that the components of image---the surface albedo, the direct illumination, and the ambient illumination---are coupled heavily in observed image. We propose to infer the shading by ordering pixels by their relative brightness, without knowing the absolute values of the image components beforehand. The pairwise shading orders are estimated in two ways: brightness order and low-order fittings of local shading field. The brightness order is a non-local measure, which can be applied to any pair of pixels including those whose reflectance and shading are both different. The low-order fittings are used for pixel pairs within local regions of smooth shading. Together, they can capture both global order structure and local variations of the shading. We propose a Consistency-aware Selective Fusion (CSF) to integrate the pairwise orders into a globally consistent order. The iterative selection process solves the conflicts between the pairwise orders obtained by different estimation methods. Inconsistent or unreliable pairwise orders will be automatically excluded from the fusion to avoid polluting the global order. Experiments on the MIT Intrinsic Image dataset show that the proposed model is effective at recovering the shading including deep shadows. Our model also works well on natural images from the IIW dataset, the UIUC Shadow dataset and the NYU-Depth dataset, where the colors of direct lights and ambient lights are quite different.
\end{abstract}

}

\maketitle

\IEEEdisplaynontitleabstractindextext

%
\IEEEpeerreviewmaketitle

\section{Introduction}
%
%

%
%
%
%

\IEEEPARstart{A}{n} image is the result of several factors, including the material reflectance, the surface's shape, the positions and the colors of the illuminants, and the camera sensor responses. Barrow and Tenenbaum \cite{Barrow_CVR78} proposed to decompose an image into intrinsic images, each of which captures a distinct aspect of the scene. The most common outputs are the shading and the reflectance. The shading captures the strength of incident illumination at each pixel, while the reflectance shows the surface albedo. The shading is widely used to reconstruct the shapes of the surfaces \cite{Zhang99}. The albedo is invariant to illumination and geometry, so it is a robust feature for object classification and image segmentation.

In this paper we aim to recover the shading and the reflectance from a single image. This is an underconstrained problem. The absolute values of the unknown variables cannot be measured directly, since they are highly coupled in observed image. Instead, we measure the relative sizes of shading over pixels to recover its essential structure, and determine their absolute values later by boundary conditions.

We regard the shading as a global ranking of the pixels in the order of dark to bright. The boundary conditions are simply that the start points are fully shadowed pixels, while the end points are fully lit ones.

\begin{figure*}
  \centering
  \includegraphics[width=1.0\linewidth]{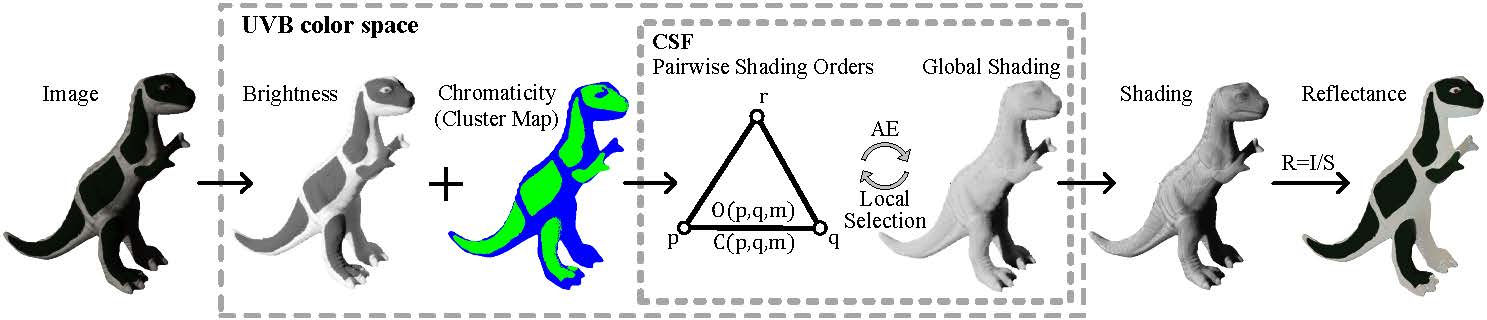}\\
  \captionsetup{font={footnotesize}}
  \caption{The flow chart of our method. Firstly the image is transformed into the $UVB$ color space. Based on brightness and cluster results over chromaticity, different methods $m$ are used to estimate the shading orders $O(p,q,m)$ between each pair of pixels $p$ and $q$. We also evaluate the reliability $C(p,q,m)$ of the estimates based on the image features. Then we use \textbf{CSF} to infer the global shading. CSF repeats two operations: \textbf{Local Selection}, \ie selecting the estimation methods and the weights for each pair of pixels under the guidance of consistency between the pairwise orders and the global shading; and \textbf{Angular Embedding (AE)}, which infers the globally consistent orders from the pairwise estimates. At last we transform the global shading back into the RGB space.}\label{fig:flow_chart}
\end{figure*}

The global shading is inferred from pairwise shading orders, which are signed differences between the shading of pixels. The flow chart is shown in Fig. \ref{fig:flow_chart}. We estimate the shading orders in the $UVB$ color space, which is spanned by a 2D shadow-free plane \cite{BIDR} and a brightness dimension. This color space has two major properties:
\begin{itemize}[leftmargin=*]
\item Pixels with the same reflectance cluster together on the shadow-free plane.
\item The brightness of image is the sum of the shading brightness and the reflectance brightness.
\end{itemize}
Based on these properties, we can use clustering-based methods to capture the global order structure of the shading. \textbf{For pixels with the same reflectance}, the shading orders can be obtained directly from the difference of the image brightness. \textbf{For pixels with different reflectance}, the shading orders can be calculated in a similar way, but the bias from the difference of the reflectance brightness should be compensated. We choose the optimal biases between different clusters of reflectance, which make the shading constant across reflectance boundaries excluding shading edges. The cluster-wise biases make it possible to handle pixel pairs whose reflectance and shading are both different.


We also model the local shading by low-order fittings to predict the shading orders between nearby pixels. Different models can capture the geometric structure of different types of surfaces. For example, a linear model can describe the shading of a smooth surface.

The estimation methods above are complementary. The clustering-based methods can be applied to any pair of pixels, in particular those of distantly located pixels, but their accuracies depend on the quality of clustering. In contrast, the low-order fittings do not rely on clustering at all, but they capture only the local structure, and the fitting errors are large for irregular surfaces.

The pairwise shading orders are combined into a global shading via Consistency-aware Selective Fusion (CSF). The major challenge is avoiding inconsistency between estimates from different methods. CSF identifies a sparse set of reliable and consistent pairwise shading orders and fuses them within a unified optimization framework. For each pair of pixels, CSF selects the most reliable estimate exclusively instead of a weighted summation of different estimates \cite{Zhao_PAMI13}\cite{Lee_ECCV12}\cite{Chen_ICCV13}. This strategy prevents unreliable estimates from polluting the results. We evaluate the reliability of pairwise orders using not only the image features but also their consistency with the global order. Therefore, the estimates that are incompatible with the majority will be suppressed, even when their preconditions happen to be satisfied by the image features. Forcing sparsity of pairwise connections further reduces unreliable measurements.

The global order is obtained from Angular Embedding (AE) \cite{AngularEmbedding}, which embeds the pixels onto a unit circle in the complex plane. AE uses a complex matrix to encode pairwise orders and their reliability simultaneously. Moreover, AE applies spectral decomposition to get a near-global optimal solution that best matches the reliable pairwise orders. After locating the darkest points on the unit circle, the absolute values of shading can be determined.

\subsection{Related Work}

Edge-based methods rely on classification of image gradients \cite{Retinex}\cite{TappenCVPR06}\cite{ShenCVPR08}. The problem is that during the integration of gradients, a single misclassified edge will result in errors in a wide area of recovered reflectance \cite{Serra_CVPR12}. Our shading orders can capture much more information than the gradients, since the objects of the measurements are no longer limited to adjacent pixels. The non-local shading orders can reduce the adverse influence of misclassified edges. Further, the long range relations define the large-scale structure directly, which can avoid the accumulation of error when integrating local measurements. Some Markov random field models \cite{Zhao_PAMI13}\cite{Lee_ECCV12}\cite{Chen_ICCV13} and dense Conditional random field models \cite{Bell_Siggraph14} also consider the relation between distant pixels. However, their non-local smooth terms are only applicable to pixels with the same reflectance or shading. This is of particular importance for pixels whose reflectance and shading are both different. Imposing shading smoothness on these pixels (\eg \cite{Bell_Siggraph14}) may cause large errors.


Solving the underconstrained problem of intrinsic image decomposition requires prior knowledge. Our method utilizes several widely used priors, including the local smoothness of shading \cite{ShenCVPR08}\cite{gehler11nips}\cite{Lee_ECCV12}\cite{Chen_ICCV13}, the piecewise constancy of reflectance \cite{ShenCVPR08}\cite{Shen_PAMI13}\cite{BarronECCV12}\cite{Chen_ICCV13}\cite{Bi2015L1Intrinsic}, and the global sparsity of reflectance \cite{gehler11nips}\cite{Shen_PAMI13}\cite{BarronECCV12}\cite{Garces2012}\cite{Nie_2014}. The local smoothness of shading guarantees the accuracy of our low-order fittings of shading. The smoothness is also the basis of inferring the cluster-wise biases of reflectance brightness. The piecewise constancy of reflectance suggests that, for a linear fitting, we can calculate the derivative of shading brightness by the derivative of image brightness. The global sparsity of reflectance ensures that the reflectance falls into a limited number of clusters, which is the basis of the clustering-based methods \cite{ShenCVPR08}\cite{Garces2012}\cite{Bell_Siggraph14}\cite{gehler11nips}\cite{Serra_CVPR12}\cite{chang14svdpgmm}. We cluster the reflectance on the data-dependent shadow-free plane derived from the Bi-illumination Dichromatic Reflection model (BIDR) \cite{BIDR}. Unlike the other shadow-free color spaces \cite{Mark92recoveringshading}\cite{ReflectanceEdge}\cite{Finlayson_eccv02}\cite{QuasiInvariant}\cite{Garces2012}, BIDR addressed explicitly the ambient illuminant that is non-negligible in natural scenes. Color Retinex adopted a degraded version of the BIDR model that predefined the normal of the shadow-free plane to be $(1,1,1)^T$ \cite{intrinsic_dataset}.

Different constraints often result in quite different shading orders. How to fuse them remains an open problem. Edge-based methods classify each edge into a reflectance edge or a shading edge. Accordingly, the shading order between the two sides of the edge can be decided. In particular, Retinex classified the edges by the magnitude of gradients \cite{Retinex}. This classification method is risky. Some shadow edges are quite strong, while the reflectance edges between similar colors are relatively weak. Extensions of Retinex introduced several new features, including texture similarity \cite{ShenCVPR08}, classifiers over local features \cite{Bell_iccv01} or patches \cite{Tappen_05}\cite{TappenCVPR06}, correlation between the mean luminance and luminance amplitude \cite{correlation}, and image sequences under different illumination directions \cite{Weiss_ICCV01}\cite{Hauagge_cvpr13}. These features improved the accuracy of classification, but none of them are robust enough to handle all kinds of scenes. CSF faces a similar problem of selecting the optimal pairwise order from several estimates. The difference is that CSF incorporates consistency between the pairwise orders and global order into the selection criteria, which can rectify the inconsistent selections made by noisy image features.

Another research stream combines different types of constraints softly by additive energies  \cite{ShenCVPR08}\cite{Shen_PAMI13}\cite{gehler11nips}\cite{Lee_ECCV12}\cite{BarronECCV12}\cite{Chen_ICCV13}\cite{Li_2014_CVPR}. This strategy avoids the risk of hard decisions faced by the edge-based methods. But this brings a new problem that different kinds of constraints are hinged together. As a result, the meta-constraints will be violated more or less. Specifically, the local smoothness of shading tends to weaken sharp shadow edges, and the piecewise constancy of reflectance will erase the texture.




Ranking elements from their pairwise comparisons has been extensively studied in many fields \cite{Agrawal_2005}\cite{AE_luminance}\cite{Maire10}. Angular Embedding \cite{AngularEmbedding} adopts a cosine error function, which is proven to be more robust to outliers than the traditional $L1$ or $L2$ errors used by Least Squares Embedding \cite{Agrawal_2005}. Angular Synchronization (AS) also uses the angular space \cite{AngularSync}, but it does not consider the confidences of pairwise measures.

Many recent methods address more intrinsic components other than shading and reflectance, including specular reflection \cite{Beigpour_ICCV11}, shape and illumination \cite{BarronECCV12}, coarse-scale and detailed shading \cite{Liao_cvpr13}, direct and indirect irradiance \cite{Chen_ICCV13}\cite{CarrollRA11}, illuminant color and sensor characteristics \cite{marc_cvpr2014}, and texture \cite{Jeon14texture}. These detailed decompositions give a more comprehensive analysis of the scene, but they also make the problem much more complex. Recently, new constraints have been formed based on the geometric information of RGB-Depth images \cite{Lee_ECCV12}\cite{Barron2013A}\cite{Chen_ICCV13}\cite{Jeon14texture}. We use the depth map to render shadow maps that can determine the positions of shading edges. More recently, the intrinsic video techniques have extended the research to videos \cite{Kong2014IntrinsicVideo}\cite{Ye2014IntrinsicVideo}\cite{Bonneel2014IntrinsicVideo}.

The main contributions of this work are: (1) we infer shading from pairwise shading orders, which are more informative and flexible than image gradients; (2) we propose a group of complementary methods to estimate the shading orders for different types of pixel pairs; (3) we adopt a strategy of local competition and global assimilation to select reliable and consistent pairwise orders and combine them into a global shading by AE; and (4) we introduce the $UVB$ color space to deal with images under considerable ambient illuminants.

This paper is an extension of our conference paper \cite{liu_accv2014}. The primary new contributions are: (1) we propose CSF to achieve local selection and global combination in a unified optimization framework in Section \ref{sec:infer_shading}; (2) we apply our model to the RGB-Depth images by incorporating new features from the depth map in Section \ref{sec:reliability}; and (3) we test on natural images from the IIW dataset in Section \ref{sec:IIW}.

\section{Image Formation}
\label{sec:IRS}

An image with only body reflection can be modeled as \cite{BIDR}
\begin{equation}
\label{eqn:BIDR}
\I^i(p) = \R_b^i(p)(\gamma(p) \L_d^i+\L_a^i),
\end{equation}
where the superscript $i$ indexes the RGB channels, and $p$ indexes the pixel. The body reflectance $\R_b$ denotes the diffuse reflection under white illumination. The three-dimensional vectors $\L_d$ and $\L_a$ are the direct illuminant and the ambient illuminant, respectively. $\gamma(p) \in [0,1]$ is the direct shading, \ie the proportion of direct illumination reaching the surface. BIDR assumes that the direct and ambient illuminants are constant across the materials \cite{BIDR}. When there are multiple direct illuminants with the same color, their effects can be added. 

Inspired by the shadow removal problem \cite{Guo_PAMI12}, we define the \textbf{reflectance} to be the image lit by full direct illuminant together with the ambient illuminant:
\begin{equation}
\label{eqn:reflectance}
\R^i(p)=\R_b^i(p)(\L_d^i+\L_a^i).
\end{equation}
Accordingly, the \textbf{shading} is defined to be
\begin{equation}
\label{eqn:shading}
\S^i(p) = \frac{\I^i(p)}{\R^i(p)} = \frac{\gamma(p) \L_d^i + \L_a^i}{\L_d^i+\L_a^i}.
\end{equation}
For a fully lit area (\ie $\gamma=1$), the shading reaches its maximum. For a fully shadowed area (\ie $\gamma(p)=0$), the shading will be $\S(p)=\L_a/(\L_d+\L_a)$. In natural scenes, the direct lights are always much stronger than the ambient lights, so the shading of fully shadowed areas should be a small positive value.

The color of the shading in (\ref{eqn:shading}) does not have definite physical meaning, so we show the shading in grayscale for all the figures in this paper following \cite{intrinsic_dataset} and \cite{Bell_Siggraph14}. Readers that are interested in the color of the shading are referred to the supplementary material for several examples.

\section{Shading Orders from Brightness}
\label{sec:LocalGamma}

We infer shading orders in the $UVB$ color space. We will show that image brightness has a linear relation to the log of shading. Therefore pairwise shading orders can be estimated by either brightness orders or low-order fittings of local shading.

\subsection{The $UVB$ Color Space}
\label{sec:brightness}

\begin{figure}
  \centering
  \includegraphics[width=1.0\linewidth]{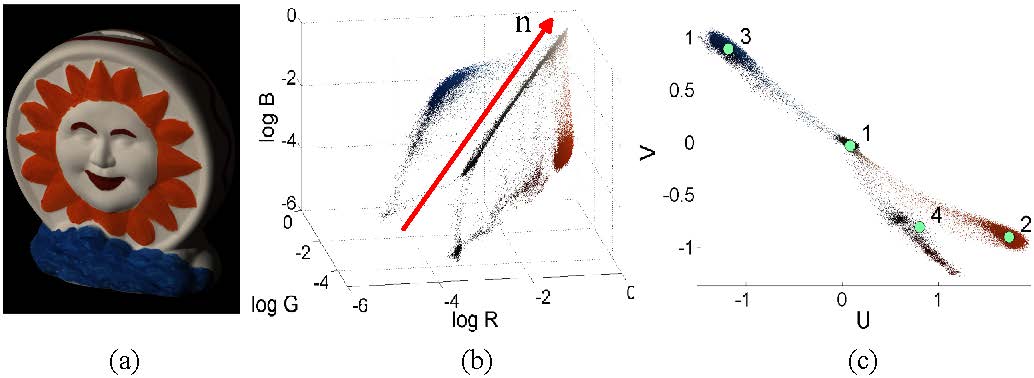}\\
  \captionsetup{font={footnotesize}}
  \caption{The brightening direction and the shadow-free plane. (a) The raw image. (b) The pixels in log RGB space. The white, orange, blue and dark red pixels form four nearly parallel lines. The arrow $\n$ indicates the brightening direction. (c) Projections of pixels on the shadow-free plane perpendicular to the brightening direction. The pixels fall into four groups on the shadow-free plane, each for a distinct color. The green dots indexed by 1$\sim$ 4 are cluster centers for white, orange, blue and dark red pixels, respectively.}\label{fig:brighten}
\end{figure}

The BIDR model delivers a 2D shadow-free plane $UV$ \cite{BIDR}. The normal $\n$ of the $UV$ plane points from the shadowed pixels to the lit ones sharing the same body reflectance $\R_b$ (see Fig. \ref{fig:brighten}b for an example). We call the normal $\n$ the \textbf{brightening direction}. Formally, the brightening direction is defined by
\begin{equation}
\small
\label{eqn:n}
\n = \frac{1}{K}\left(\log\I(p)|_{\gamma(p)=1}-\log\I(q)|_{\gamma(q)=0}\right)=\frac{1}{K}\log(\frac{\L_d}{\L_a}+1),
\end{equation}
where the pixels $p$ and $q$ should satisfy $\R_b(p)=\R_b(q)$, and $K$ is the normalization factor.

From (\ref{eqn:n}) we can see that the brightening direction depends only on the ratio of illuminants, so all the pixels share the same brightening direction (Fig. \ref{fig:brighten}b). If the ratio of illuminants is unknown, we can search the most probable brightening direction that minimizes the entropy of pixels on the $UV$ plane \cite{BIDR}\cite{Finlayson09}. Since pixels with similar reflectance $\R_b$ will stay closely together on the $UV$ plane (Fig. \ref{fig:brighten}c), the entropy of the distribution of pixels will be minimized.

Let $\u$ and $\v$ be any pair of basis vectors on the $UV$ plane. Then we have a rotation matrix $H=[\u,\v,\n]$ that transforms the $\log$ RGB space into a new color space $UVB$:
\begin{equation}
\small
\label{eqn:UVB}
[I^u(p),I^v(p),I^b(p)]=\log\I(p)H.
\end{equation}
The dimension $I^b$ captures the intensity of the image, and we call it the \textbf{brightness}.

According to (\ref{eqn:shading}) and (\ref{eqn:UVB}), the brightness of the image can be factorized as follows:
\begin{equation}
\small
\label{eqn:Iw}
I^b(p)= \log \S(p) \cdot \n + \log \R(p) \cdot \n = S^b(p) + R^b(p).
\end{equation}
Here we used the fact that $\log\I(p)=\log\R(p)+\log\S(p)$. The \textbf{shading brightness} is $S^b(p)= \log \S(p) \cdot \n$, which is a linear function of $\log\S$. The \textbf{reflectance brightness} $R^b(p)=\log \R(p) \cdot \n$ can be regarded as a bias determined by the body reflectance $\R_b$. This linear relationship is the basis for estimating the shading orders in Section \ref{sec:shading_order}.

According to (\ref{eqn:UVB}), the shading in the $UVB$ space should be $[S^u(p), S^v(p), S^b(p)]=\log\S(p) H$. Note that $S^u$ and $S^v$ are nearly zero since the $UV$ plane is shadow-free \cite{BIDR}. \textbf{The only unknown dimension is the shading brightness $S^b$}, and we will infer it from pairwise shading orders in Section \ref{sec:infer_shading}.

Once we obtain $S^b$, the shading in $RGB$ space can be recovered by
\begin{equation}
\small
\label{eqn:recover_S}
\S(p) = exp([S^u(p),S^v(p),S^b(p)]H^{-1}),
\end{equation}
where $exp$ denotes element-wise exponential. Note that the rotation matrix $H$ is always invertible. The reflectance can be obtained from $\R(p)=\I(p)/\S(p)$.

\subsection{Measuring Pairwise Shading Orders}
\label{sec:shading_order}

\begin{figure}
  \centering
  \includegraphics[width=1.0\linewidth]{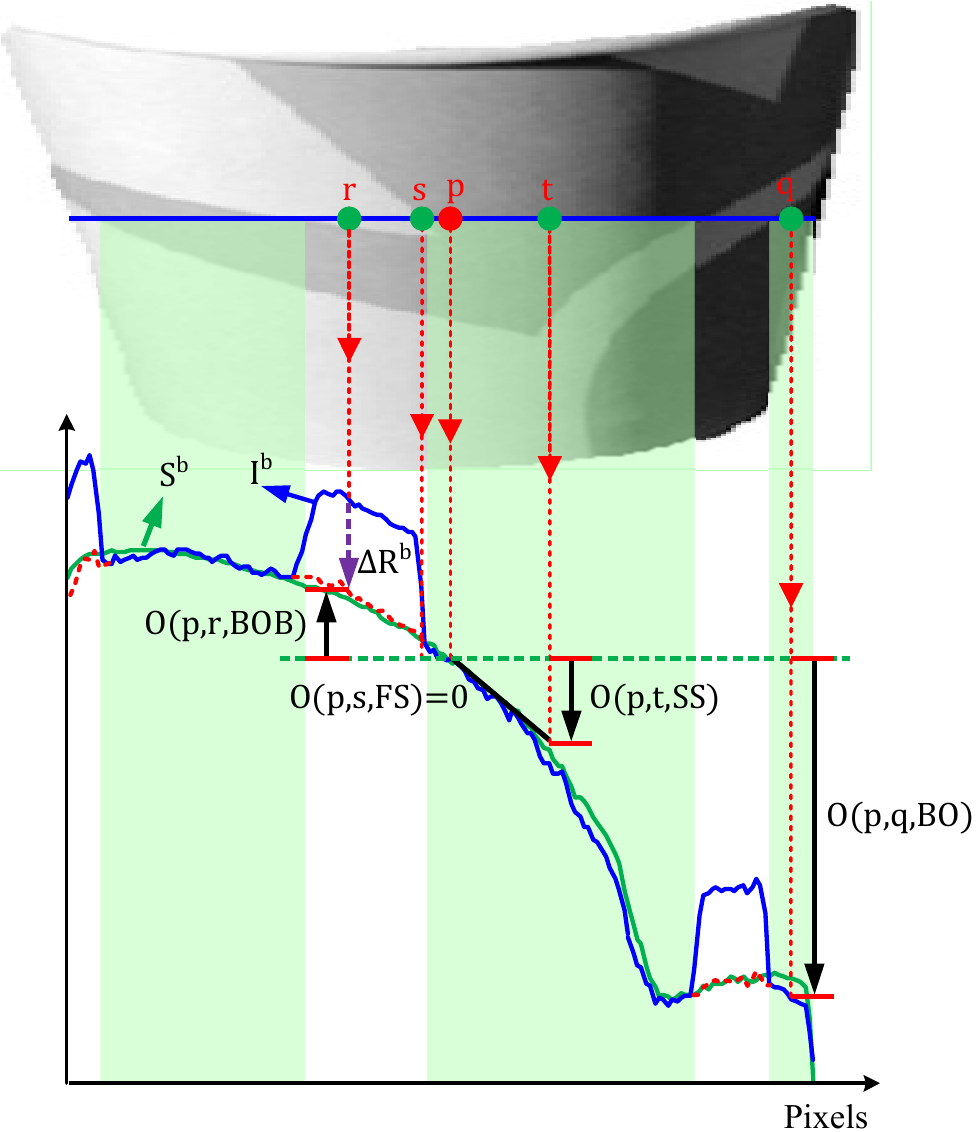}\\
  \captionsetup{font={footnotesize}}
  \caption{Calculating shading orders $O$ from brightness $I^b$. We align the curves of the brightness $I^b$ and the ground-truth shading brightness $S^b$ to make $I^b(p)=S^b(p)$. The red dashed curve is the brightness after compensating the bias of reflectance brightness $\Delta R^b$. The green masks cover the green pixels while the uncovered ones are white.}\label{fig:shading_order}
\end{figure}

The \textbf{shading order} between pixels $p$ and $q$ is defined to be the signed difference between shading brightnesses, \ie $O(p,q)= S^b(p)-S^b(q)$. We propose four methods $\mathcal{M}=\{BO,BOB,FS,SS\}$ to estimate the shading orders. These methods are shown in Fig. \ref{fig:shading_order}.

\noindent\textbf{Brightness Order (BO).} According to (\ref{eqn:Iw}), if two pixels have the same reflectance brightness $R^b$ or equivalently, the same body reflectance $\R_b$, their shading order will be equal to their difference of brightnesses:
\begin{equation}
\small
\label{eqn:BO}
O(p,q,BO)= I^b(p)-I^b(q) \ \ \mbox{if } \R_b(p) = \R_b(q).
\end{equation}

\noindent\textbf{Brightness Order minus Bias (BOB).} For pixels with different body reflectance, the bias of reflectance brightness $\Delta R^b$ should be compensated as follows:
\begin{equation}
\small
\label{eqn:BOB}
O(p,r,BOB)= I^b(p)-I^b(r) - \Delta R^b(p,r) \  \mbox{if } \R_b(p) \neq \R_b(r),
\end{equation}
where $\Delta R^b(p,r)=R^b(p)-R^b(r)$ is the bias. The process of calculating the bias will be described in Section \ref{sec:Clustering}. BO and BOB together can estimate the shading order between any two pixels.

For pixels nearby, we can fit their shading brightness by low-order functions. This is based on the assumption of local smoothness of shading, which is valid for most parts of natural images.

\noindent\textbf{First-order Smoothness (FS).} For flat surfaces, the normal directions and thus the incident angles change little. According to the cosine law of the Lambertian reflection, the variation of shading brightness will be small. The first-order derivative of shading brightness should be almost zero if there are no shadow edges. Consequently, the adjacent pixels will have nearly identical shading brightness:
\begin{equation}
\small
\label{eqn:FS}
O(p,s,FS)= 0  \ \ \mbox{if }  s\in \mathcal{N}(p),\frac{\partial I^b(p)}{\partial p} \approx 0,
\end{equation}
where $\mathcal{N}(p)$ is the neighborhood of $p$, and $\frac{\partial I^b(p)}{\partial p}$ is the derivative of $I^b$ evaluated at $p$.

\noindent\textbf{Second-order Smoothness (SS).} For smooth surfaces, the surface normal rotates smoothly. As a result, the shading brightness will change smoothly. We assume that the second-order derivative of the shading is close to zero. Thus we can fit the local shading by a linear function. We further assume that the adjacent pixels share the same body reflectance, so the slope of the linear model $\frac{\partial S^b(p)}{\partial p}=\frac{\partial I^b(p)}{\partial p}$. The shading order between two nearby pixels will be
\begin{equation}
\small
\label{eqn:SS}
O(p,t,SS)= \frac{\partial I^b(p)}{\partial p} \cdot  (p-t) \ \ \mbox{if } t\in \mathcal{N}(p), \frac{\partial^2 (I^b(p))}{\partial p ^2} \approx 0,
\end{equation}
where $p-t$ is the directed spatial distance between $p$ and $t$. In practice, we calculate the derivative and the spatial distance in the horizontal and vertical directions separately.

The preconditions of the methods above are not mutually exclusive, so different methods may be applicable to the same pair of pixels. The preconditions together cover all possible situations, so we can find at least one suitable method for most pairs of pixels. The redundancy and completeness of these methods are the basis for robust estimates of shading orders.

The biases of reflectance brightness $\Delta R^b$ in (\ref{eqn:BOB}) are needed to estimate the shading orders between pixels with different body reflectance.

For each pair of pixels, we obtained several estimates of their shading order by different methods in Section \ref{sec:shading_order}. These methods rely on certain assumptions about the scene, which may be invalid for certain parts of the image. Therefore, the estimated shading orders may differ from the ground-truth. We evaluate the reliability of each estimate by checking whether influential perturbations happened there.

\subsection{Estimating the Bias of Reflectance Brightness}
\label{sec:Clustering}

The biases of reflectance brightness $\Delta R^b$ in (\ref{eqn:BOB}) are needed to estimate the shading orders between pixels with different body reflectance. The absolute values of reflectance brightness $R^b$ are unavailable, so we cannot calculate their biases directly. Instead, we cluster the pixels by body reflectance, and estimate the biases of reflectance brightness between different clusters. The local smoothness of shading implies that pixels within a small patch have similar shading brightness. According to (\ref{eqn:Iw}), the bias of reflectance brightness between two clusters can be approximated by their difference of image brightness within small patches.

\begin{figure}
  \centering
  \includegraphics[width=1.0\linewidth]{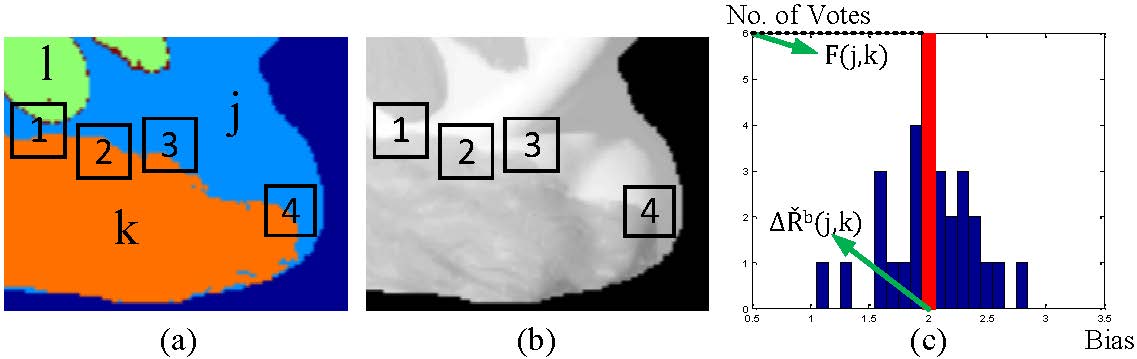}\\
  \captionsetup{font={footnotesize}}
  \caption{Estimating the bias of reflectance brightness between reflectance categories. (a) The cluster map. The symbols $j$, $k$, and $l$ stand for 3 reflectance categories. The squares indicate representative patches for estimating the bias of reflectance brightness between categories $j$ and $k$; (b) The brightness $I^b$. The biases obtained from patches 3 and 4 are outliers, since there are shadow edges inside them; (c) The histogram of the patch-wise biases of reflectance brightness between categories $j$ and $k$. The peak of the histogram is selected to be the result.}
  \label{fig:Ib}
\end{figure}

The main process is shown in Fig. \ref{fig:Ib}. The image is divided into dense grids with 10 pixels in each side. For a patch $T$ containing pixels from both categories $j$ and $k$, the difference of reflectance brightness is calculated by $\Delta R^b(j,k,T)=\bar{I}^b(j,T)-\bar{I}^b(k,T)$, where $\bar{I}^b(j,T)$ and $\bar{I}^b(k,T)$ are the median brightness of pixels belonging to categories $j$ and $k$, respectively. We generate a histogram of the patch-wise measures $\Delta R^b(j,k,T)$, and take the highest peak to be the estimate $\Delta \check{R}^b(j,k)$ as shown in Fig. \ref{fig:Ib}c. The minorities of the histogram mainly come from patches with shading edges in them (\eg patches 3 and 4 in Fig. \ref{fig:Ib}b). The reliability $F$ of the estimate is set to be the number of votes from the patches. When $F_{j,k}$ is 0, it means categories $j$ and $k$ are not adjacent, and their bias cannot be measured directly. In this case, we resort to their biases with other categories.

Taking each reflectance category as a node, we can build an undirected graph $\mathcal{G}=(\mathcal{V}, \mathcal{E})$, where $\mathcal{V}$ is the set of nodes and $\mathcal{E}$ is the set of edges. The weight of the edge between node $j$ and $k$ is set to be $1/F_{j,k}$, where $F_{j,k}$ is the reliability of $\Delta \check{R}^b(j,k)$ as described before. We can get an estimate of the bias between two nodes by summing the biases along any path connecting them. We further eliminate the multi-path effect by extracting the Minimum Spanning Tree (MST) of the graph $\mathcal{G}$. The MST ensures that there is one and only one path between any two nodes, so the relative reflectance brightness $\check{R}^b$ of each node can be uniquely determined. Meanwhile, the total reliability of the remaining pairwise biases is maximized.

The sparsity of the reflectance spectra \cite{color_lines} ensures that the pixels can be clustered into a small number of categories. Since pixels on the shadow-free plane $UV$ are well organized by their body reflectance, we cluster the pixels by a simple k-means. The number of clusters is set to be the number of local maxima in the 2D histogram of $I^u$ and $I^v$. The bin size of the histogram is empirically set to be 0.03.

\section{The Reliability of Pairwise Orders}
\label{sec:reliability}

For each pair of pixels, we obtained several estimates of their shading order by different methods in Section \ref{sec:shading_order}. These methods rely on certain assumptions about the scene, which may be invalid for certain parts of the image. Therefore, the estimated shading orders may differ from the ground-truth. We evaluate the reliability of each estimate by checking whether influential perturbations happened there.

The \textbf{reliability} of an estimate is the probability of all its premises being valid, which is calculated by a Noisy-Or model
\begin{equation}
\small
\label{eqn:confidence}
C(p,q,m) = \prod_{f\in \mathcal{C}_m} 1-P_{f}(p,q),\ \  m\in \mathcal{M},
\end{equation}
where $\mathcal{C}_m$ is the set of perturbations that the method $m$ is not robust to, as listed in Table \ref{tab:reliability}. The probability $P_{f}(p,q)$ measures how likely the perturbation $f$ occurs around pixels $p$ and $q$. For an ideal image without any perturbation, all the methods get equally high confidences. Once a perturbation happens, the confidences of sensitive methods will drop.

\begin{table}
\captionsetup{font={footnotesize}}
\caption{The robustness of the estimation methods with respect to different perturbations.}
\label{tab:reliability}
\centering
\begin{tabular}{l|c|c|c|c} \hline
Perturbations                           & BO            & BOB           & FS                      & SS \\ \hline
Clustering Error (CE)                   & Yes           & No            & Yes                     & Yes     \\
Local Color Variance (LCV)              & No            & No            & Yes                     & No      \\
Shadow Edges (SE)                       & Yes           & Yes           & No                      & No      \\
Reflectance Change (RC)                 & No            & Yes           & Yes                     & Yes     \\
Surface Normal Change (SNC)             & Yes           & Yes           & No                      & Yes     \\
Spatial Distance (SD)                   & Yes           & Yes           & No                      & Moderate \\ \hline
\end{tabular}
\end{table}

The occurrences of the perturbations are predicted by image features. Generally, we calculate a distance $x$ between the pair of pixels according to each feature, and translate the distance into probability by a sigmoid function in the form of $sigm(x;w)=\frac{2}{1+e^{-wx}}-1$, where $w$ is a positive weight. The features are described below.

\begin{figure}
  \centering
  \includegraphics[width=1.0\linewidth]{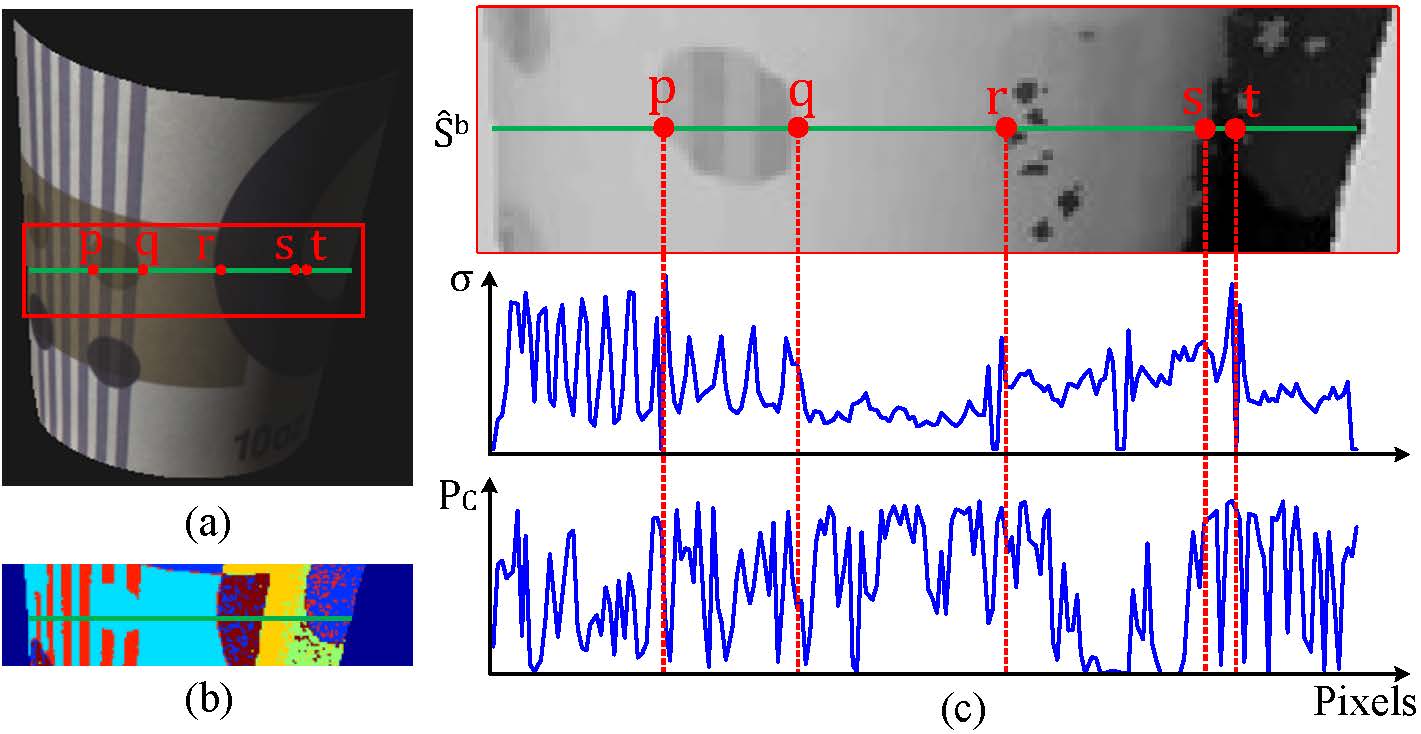}\\
  \captionsetup{font={footnotesize}}
  \caption{Basic features for calculating CE and LCV. (a) The raw image; (b) The cluster map; (c) The shifted shading brightness $\hat{S}^b$, the color variation $\sigma$, and the cluster probabilities $P_C$.}
  \label{fig:perturbation}
\end{figure}

\noindent\textbf{Clustering Error (CE)} is the probability that the clustering of pixels on the shadow-free plane is inaccurate, which is calculated by
\begin{equation}
\small
\label{eqn:CE}
P_{CE}(p,q) = \left(1-P_C(p)P_C(q)\right)\cdot sigm(e_{\hat{S}^b}(p,q);w_{1}),
\end{equation}
where the cluster probability $P_C$ is the likelihood of each pixel belonging to its reflectance category, and $e_{\hat{S}^b}$ is the strength of the step edge \cite{ncut} on the shifted shading brightness $\hat{S}^b$. The first term increases as the pixel $p$ or $q$ deviates from the cluster centers. The second term is large when the pixels are improperly categorized or the relative reflectance brightnesses are inaccurately estimated, as shown in Fig. \ref{fig:perturbation}c. Here each reflectance category is modeled by a multivariate normal distribution. The shifted shading brightness $\hat{S}^b$ is obtained from the brightness $I^b$ minus the relative reflectance brightness $\check{R}^b$ (Section \ref{sec:Clustering}) followed by a median filtering.

\noindent\textbf{Local Color Variance (LCV)} is defined to be:
\begin{equation}
\small
\label{eqn:LCV}
P_{LCV}(p,q) = sigm(\max(\sigma(\I(p)), \sigma(\I(q)));w_{2}),
\end{equation}
where $\sigma(\I(p))$ is the standard deviation of chromaticities $I^u$ and $I^v$ within the 3x3 window centered at pixel $p$. Large color variations mainly appear at reflectance boundaries (Figs. \ref{fig:perturbation}a and \ref{fig:perturbation}c). 

\noindent\textbf{Shadow Edges (SE)} are caused by occlusions of the direct light. To locate the shadow edges, we render the direct shading $\tilde{\gamma}$ under uniformly sampled illuminants. The direct shading is similar to the visibility map proposed by Lee \etal \cite{Lee_ECCV12}. The difference is that they assume the illuminants to be infinitely far away, which is inaccurate for indoor scenes. Instead, we sample the feasible positions of the illuminant within the room box. The probability of a shadow edge between pixels $p$ and $q$ is calculated by their direct shading under promising illuminants, as follows:
\begin{equation}
\small
\label{eqn:SE}
P_{SE}(p,q) = sigm(\frac{1}{|\mathcal{L}|}\sum_{\L_d\in \mathcal{L}}\|\tilde{\gamma}(\L_d,p)-\tilde{\gamma}(\L_d,q)\|;w_{3}).
\end{equation}
Here $\mathcal{L}$ is the set of promising illuminants, and $\tilde{\gamma}(\L_d,p)$ is the direct shading at pixel $p$ under illuminant $\L_d$. We select the promising illuminants according to the correlation between the rendered direct shading $\tilde{\gamma}$ and the brightness $I^b$. See the supplementary material for details. The Shadow Edges feature is not applicable to RGB-only images, since the geometric layout is needed for rendering the shading map.


\noindent\textbf{Reflectance Change (RC)} distinguishes pixels with different chromaticities or intensities, which are assumed to have different reflectance \cite{intrinsic_dataset}\cite{gehler11nips}\cite{Garces2012}\cite{Bell_Siggraph14}. We calculate the probability of a reflectance change by
\begin{equation}
\small
\label{eqn:RC}
P_{RC}(p,q) = sigm(d_{uv}(p,q);w_{4})\cdot sigm(e_{b}(p,q);w_{5}), 
\end{equation}
where $d_{uv}$ is the geometric distance on the shadow-free plane. $e_b(p,q)$ is the magnitude of the step edge lying between $p$ and $q$ in the brightness $I^b$, which aims at distinguishing colors with similar chromaticity but different intensities, especially achromatic ones.

\noindent\textbf{Surface Normal Change (SNC)} generates shading variation \cite{Lee_ECCV12}\cite{Chen_ICCV13}\cite{Jeon14texture}. We calculate the probability of a surface normal change by
\begin{equation}
\small
\label{eqn:SNC}
P_{SNC}(p,q) =  sigm(\angle(\mathbf{N}(p), \mathbf{N}(q));w_{6}),
\end{equation}
where $\angle(\mathbf{N}(p), \mathbf{N}(q))$ is the angle between the surface normals at pixels $p$ and $q$. The surface normals are derived from the depth map \cite{Lee_ECCV12}. SNC is unavailable to RGB-only images.

\noindent\textbf{Spatial Distance (SD)} is simply the geometric distance between the pixels \cite{Chen_ICCV13}\cite{Bell_Siggraph14}:
\begin{equation}
\small
\label{eqn:SD}
P_{SD}(p,q) = sigm(d_s(p,q);w_{7}).
\end{equation}
For RGB-Depth images, we first calculate the 3D positions of the pixels in camera coordinates and then compute their distances. For RGB-only images, we use the 2D coordinates in the image plane.

\textbf{Discussion.} The features above can help us choose the best estimation method for a certain pair of pixels. Among them, CE focuses on whether the biases of reflectance brightnesses are correctly estimated, which is the key to the success of the BOB method. We check the correctness by both the cause and the effect, \ie the pixels are tightly clustered and the estimated shading is smooth, respectively.

LCV and RC capture the local and large-scale behaviour of reflectance change, respectively. The local variation, coupled with image blur, will disturb the measurements of brightness as well as its gradient. This will cause problems in most estimation methods except for the FS, which is only concerned with the adjacency of pixels.




\section{Global Shading from Shading Orders via Consistency-aware Selective Fusion}
\label{sec:infer_shading}

Thus far we have obtained a matrix $O$ of the pairwise shading orders (Section \ref{sec:shading_order}) together with a confidence matrix $C$ from (\ref{eqn:confidence}) representing their reliability. Now we use the Consistency-aware Selective Fusion (CSF) to select a subset of reliable and consistent pairwise orders, and combine them into an optimal global order. CSF is designed under the following criteria:
\begin{itemize}[leftmargin=*]
\item For a pair of pixels $p$ and $q$, the optimal estimation method $M_{p,q}\in \mathcal{M}$ is selected exclusively.
\item The pairwise connections $W_{p,q}$ should be sparse such that the outliers are excluded.
\item The total confidence of the selected pairwise shading orders should be maximized.
\item The global order should match the input pairwise orders.
\end{itemize}

\begin{figure}
  \centering
  \includegraphics[width=1.0\linewidth]{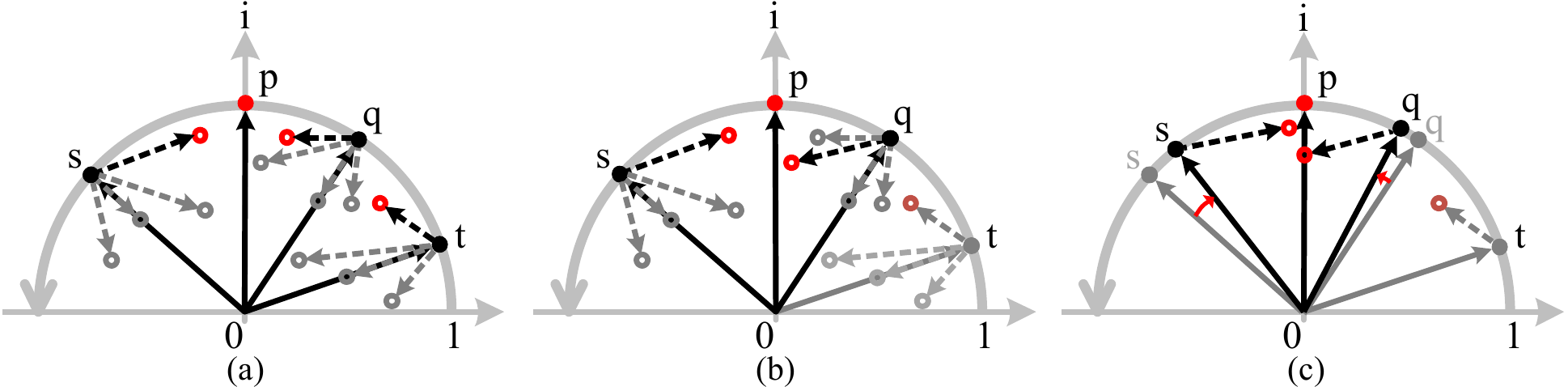}
  \captionsetup{font={footnotesize}}
  \caption{An illustration of CSF. (a) Initial selection of estimation methods $M$ based on the confidence only; (b) Update estimation methods $M$ and pairwise connections $W$ by the consistency of the pairwise orders to the global order; (c) Updated embedding $Z$. Each neighboring pixel will generate four different estimates of the position of $p$, indicated by rings. The red rings are those being selected for AE.}
  \label{fig:CSF}
\end{figure}

In practice, the global order is obtained through Angular Embedding (AE) \cite{AngularEmbedding}. Let $Z_p=e^{iS^b(p)}$ with $i=\sqrt{-1}$ denote the embedding of pixel $p$ on the unit circle in the complex plane (Fig. \ref{fig:CSF}). The angle $\Theta_{p,q}$ from $Z_p$ to $Z_q$ is the shading order between $p$ and $q$. AE finds an embedding that makes $\Theta_{p,q}$ consistent with the input shading order $O_{p,q}=O(p,q,M_{p,q})$.

The estimation methods $W$, the pairwise connections $M$ and the embedding $Z$ are optimized jointly as follows:
\begin{equation}
\label{eqn:optimization}
\begin{aligned}
&\min_{W,M,Z} J_{AE}(Z;W,M) + P(W)\\
&\mbox{s.t. } \|Z_p\|=1,\ \sum_q C_{p,q}=D_p,\forall p,\  W(p,q)\geq 0,\forall p,q,\\
\end{aligned}
\end{equation}
where the errors of Angular Embedding is defined to be \cite{AngularEmbedding}
\begin{equation}
\label{eqn:angular_embedding}
J_{AE}(Z;W,M)= \sum_{p,q}C_{p,q}\cdot \|Z_p-Z_q e^{iO_{p,q}}\|^2,
\end{equation}
and the regularization term is in the form of elastic net \cite{Zou05enet}
\begin{equation}
\label{eqn:elastic_net}
P(W)=\alpha_1\|W\|_1 + \frac{\alpha_2}{2}\|W\|^2_2.
\end{equation}
Here $C_{p,q}=W_{p,q}C(p,q,M_{p,q})$ is the weighted confidence. The diagonal matrix $D_p= \sum_{q} \max_{m\in \mathcal{M}} C(p,q,m)$ is a degree matrix. $\alpha_1$ and $\alpha_2$ are the weights of lasso ($L1$) and ridge ($L2$), respectively. Elastic net enforces group sparsity on the weights, so several groups of reliable neighbors will be selected for each pixel.

We optimize the variables $M$, $W$, and $Z$ iteratively as described in Algorithm \ref{alg:optimization}. Fig. \ref{fig:CSF} illustrates one iteration of the process. The details are given below.

\begin{algorithm}[t!]
\caption{Consistency-aware Selective Fusion}
\label{alg:optimization}
\begin{algorithmic}[l]
\REQUIRE
Pairwise shading orders $O$ and the relative confidence $C$, the initial weights $\alpha_1$ and $\alpha_2$ of regularizer, the threshold $\omega_{min}$ of the density of non-zero elements in $W$, and the step size $\tau$.
\ENSURE
Embedding $Z$.
\noindent \STATE Initialization: $W=1_{n,n}$, where $n$ is the number of pixels; $M_{p,q}=\arg\max_m C(p,q,m)$.
\WHILE {$\alpha_2>0$}
        \STATE Optimize $Z$ using (\ref{eqn:angular_embedding});
        \STATE Choose $M$ using (\ref{eqn:opt_M});
        \STATE $\alpha_2=\alpha_2-\tau$;
        \STATE Update $W$ using (\ref{eqn:opt_W});
        \IF {$\|W\|_0<\omega_{min} n^2$}
        \STATE Break;
        \ENDIF
\ENDWHILE
\RETURN $Z$.
\end{algorithmic}
\end{algorithm}

\noindent\textbf{Choose $M$}. Keeping $W$ and $Z$ fixed, we can search the optimal estimation method by
\begin{equation}
\label{eqn:opt_M}
\begin{aligned}
&\arg\min_M \sum_{p,q}W_{p,q}C(p,q,M_{p,q})\cdot \|Z_p-Z_q e^{iO(p,q,M_{p,q})}\|^2 \\
&\mbox{s.t. } \sum_q W_{p,q}C(p,q,M_{p,q})=D_p,\forall p.
\end{aligned}
\end{equation}
It can be optimized by the Lagrange method. We iteratively pick the optimal $M_{p,q}$ that balances the confidence and the consistency of orders under the current Lagrangian multiplier, and updates the multiplier by dual ascent. In Fig. \ref{fig:CSF}b the selected method for pixels $p$ and $q$ is the one with the second highest confidence but the best consistency to the global shading.

\noindent\textbf{Update $W$}. Keeping $M$ and $Z$ fixed, we can update $W$ by
\begin{equation}
\label{eqn:opt_W}
\begin{aligned}
&\arg\min_W \sum_{p,q}W_{p,q} E_{p,q} + \alpha_1\|W\|_1 + \frac{\alpha_2}{2}\|W\|^2_2\\
&\mbox{s.t. } \sum_q W_{p,q}\tilde{C}_{p,q}=D_p,\forall p,\ \  W_{p,q}\geq 0,\forall p,q, \\
\end{aligned}
\end{equation}
where $\tilde{C}_{p,q}=C(p,q,M_{p,q})$ and the confidence-weighted embedding error is $E_{p,q}=\tilde{C}_{p,q}\cdot\|Z_p-Z_qe^{iO(p,q,M_{p,q})}\|^2$. This optimization problem can be solved by the Alternating Direction Method of Multipliers (ADMM) \cite{ADMM}. See the supplementary material for details.

From (\ref{eqn:opt_W}) we can see that the larger the embedding error $E(p,q)$ is, the smaller $W(p,q)$ tends to be. This can be observed in Fig. \ref{fig:CSF} that the pair of $p$ and $t$ gets a low weight, since the embedding error is large for every estimation method. Note that we decrease the value of $\alpha_2$ gradually in Algorithm \ref{alg:optimization}, which makes $W$ more and more sparse. This progressive sparsity has better numerical stability than setting $\alpha_2$ to be a small value in the very beginning. As $\alpha_2$ gets too small, the pairwise connections may become overly sparse, producing an ill-conditioned graph. We terminate the iteration of Algorithm \ref{alg:optimization} in this case.

\noindent\textbf{Optimize $Z$}. Optimizing the embedding error $J(Z;W,M)$ in (\ref{eqn:angular_embedding}) directly is hard in practice since it has $n$ constraints, where $n$ is the number of pixels. Relaxing the unit-length constraints in (\ref{eqn:optimization}) to be $Z'DZ=1_n'D1_n$, the problem can be rewritten into the following matrix form:
\begin{equation}
\label{eqn:AE_matrix}
\begin{aligned}
&\min_Z Z'LZ \\
&\mbox{s.t. } Z'DZ=1_n'D1_n.
\end{aligned}
\end{equation}
Here $L$ is a Laplacian matrix
\begin{equation}
\label{eqn:L}
L = D - \left(C\bullet e^{iO} + (C\bullet e^{iO})'\right),
\end{equation}
where $\bullet$ is the matrix Hadamard product, $'$ is the complex conjugate transpose, $1_n$ is a $n \times 1$ vector of all ones, and exponentiation acts element-wise. To make the optimization tractable, we consider only the shading orders between nearby pixels, while the confidences of the other shading orders are set to be zero. In our experiments we set the neighborhood to be a square of 30 pixels in each side.

\begin{figure}
  \centering
  \includegraphics[width=1.0\linewidth]{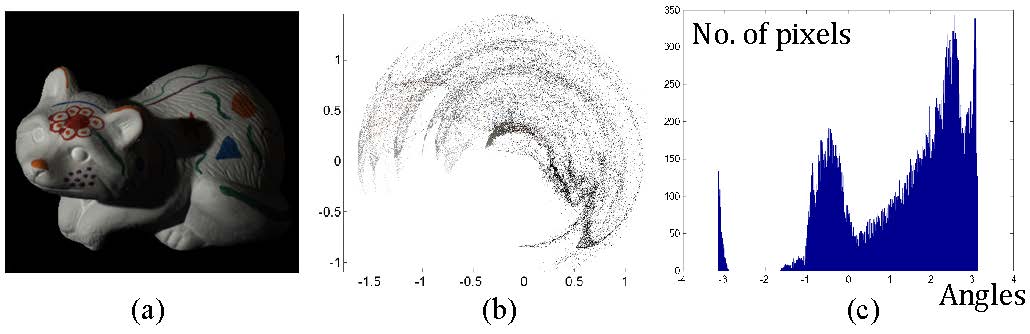}
  \captionsetup{font={footnotesize}}
  \caption{An example of the Angular Embedding result. (a) The image; (b) The output embedding; (c) The histogram of the angles of embedding.}
  \label{fig:AE}
\end{figure}

The optimization problem in (\ref{eqn:AE_matrix}) is solved by the spectral partitioning algorithm \cite{ncut} with complex-valued eigenvectors. The solution is the angles of the first eigenvector $\angle Z_0$ that has the smallest eigenvalue. We refer to the paper of Yu \cite{AngularEmbedding} for more details.

\begin{figure}
  \centering
  \includegraphics[width=1.0\linewidth]{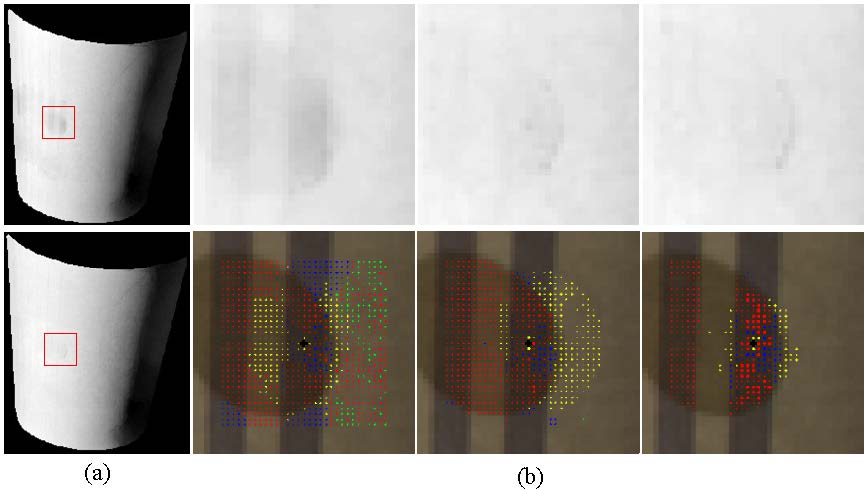}
  \captionsetup{font={footnotesize}}
  \caption{The change of variables within CSF. (a) The angles of initial embedding (Top) and the final embedding (Bottom). (b) From left to right: the states of the patch in the red box after 1, 3 and 7 iterations, respectively. Top row: the angles of embedding; Bottom row: the selected methods to estimate the shading orders from the central pixel (marked with a black cross) to its neighbors. Red: BO; Green: BOB; Yellow: FS; Blue: SS. The sizes of the dots indicate the weighted confidences of the pairwise orders. The original image is shown in Fig. \ref{fig:perturbation}a.}
  \label{fig:iteration}
\end{figure}

\textbf{Recover shading $S^b$}. To decode the shading brightness $S^b$ from the angles $\angle Z_0$, we need to ensure that the angle between any two points is less than $2\pi$, otherwise the points may overlap with each other. To achieve this, we scale the brightness dimension of the $UVB$ color space by a positive scalar. The scaling will not disturb the order of $\angle Z_0$, and we can scale the shading brightness back after the decoding.

AE allows the points to rotate as a whole around the original point. We need to rotate the points back until the angles of the darkest points are zero. Note that the darkest pixels and the brightest pixels are always separated by a gap on the circles in the complex plane. Fig. \ref{fig:AE}b shows an example. The gap can be easily located by the consecutive empty bins of the histogram of the angles $\angle Z_0$ (Fig. \ref{fig:AE}c). The pixels falling into the bins to the left of the gap are shifted to the right by $2\pi$.

Fig. \ref{fig:iteration} shows the change of variables during the iterations of CSF. In the beginning, the relative shading of some local regions are inaccurate (\eg the circle inside the red box), since some wrong estimates occasionally get higher confidences than the right ones based solely on the image features. For example, the orders obtained from the BOB method (indicated by green dots) may possibly be wrong since the clustering is inaccurate (see Fig. \ref{fig:perturbation}b). Some pixels with similar but different colors are mistaken to have the same reflectance (the red dots in the light yellow regions). Furthermore, the FS method is adopted to estimate the shading orders between distant pixels (the yellow dots far away from the center point). When the global order is used to guide the selection, the right estimation methods gradually emerge. At the same time, the weights of unreliable connections are greatly decreased as the sparsity gets stronger. Specifically, pairs of pixels whose orders cannot be accurately estimated by any method will be assigned zero weights and excluded from the fusion. As a result, the errors of $\angle Z_0$ are reduced considerably.

\section{Experiments}

We evaluate our method on the MIT Intrinsic Images dataset \cite{intrinsic_dataset}, which is a widely used benchmark. It contains ground-truth intrinsic images of 20 natural objects, and 16 of them are used for test. The images are taken in a controlled environment, where the direct illuminants are nearly white and the ambient illuminants are limited. To validate against real-world scenes, we evaluate our method on the Intrinsic Image in the Wild (IIW) dataset \cite{Bell_Siggraph14}, which is a large-scale dataset of public photo collections. We also test our method on outdoor scenes from the UIUC shadow dataset \cite{Guo_PAMI12}. We further test the utility of depth information on the RGB-Depth images from the NYU-Depth V2 dataset \cite{Silberman_ECCV12}.

\subsection{Error Metrics and Parameter Settings}

We evaluate the results on the MIT Intrinsic Images dataset primarily by the standard metric, namely the Local Mean Squared Error (LMSE) \cite{intrinsic_dataset}. However, as pointed out by Jiang \etal, LMSE is sensitive to the window size and the difference between the mean values of the recovered intrinsic images and the groundtruth \cite{correlation}. Moreover, LMSE biased towards edge-based methods \cite{Serra_CVPR12}. To give a more complete evaluation, we include the absolute LMSE (aLMSE) and the correlation metrics proposed by Jiang \etal \cite{correlation} as well as the standard MSE metric. The aLMSE is defined as follows:
\begin{equation}
aLMSE(I,\tilde{I})=\sum_w \min_a \|(I^w-\mu^w)-a(\tilde{I}^w-\tilde{\mu}^w)\|^2,
\end{equation}
where $I$ and $\tilde{I}$ are the ground-truth and estimate of intrinsic image, respectively. $w$ is the index of sliding window. $\mu$ and $\tilde{\mu}$ are the average of $I$ and $\tilde{I}$, respectively. The optimal scale $a$ is searched to minimize the square error. The influence of the difference of mean values can be eliminated by aLMSE.

The correlation is defined to be
\begin{equation}
Cor(I,\tilde{I})=\frac{E[(I-\mu)(\tilde{I}-\tilde{\mu})]}{\sigma\ \tilde{\sigma}},
\end{equation}
where $\sigma$ is the standard deviation of the image. $E$ is the expectation. We refer to the supplementary material of Reference \cite{correlation} for more details of aLMSE and correlation. Among these metrics, correlation and MSE measure the error in a global way, while LMSE and aLMSE take an average of local errors on small image windows.

For each image, the performance of reflectance and shading are calculated separately and the average of them is taken to be the result. The final result is the average of the performances over all images.


Results on the IIW dataset are evaluated by the metric of "weighted human disagreement rate" ($WHDR_{10\%}$) \cite{Bell_Siggraph14}. It measures the correct rate of judgements on "which one has a darker reflectance" between two pixels.

The main parameters of our model are the positive weights of the sigmoid function in Section \ref{sec:reliability}. We set $w_{1}$ to be $ln3/0.1$, so the sigmoid function maps a step edge of strength $0.1$ to a probability of 0.5. Similarly, we set $w_{2}\sim w_{6}$ to be $ln3/0.2$, $ln3/0.01$, $ln3/0.08$, $ln3/0.1$, and $ln3/0.2$, respectively. Specifically, we set the $w_{7}$ of the FS method to be twice as much as that of the SS method. We find the medium of the spatial distances of all the pixel pairs $\bar{d}_s$, and set $w_{7}$ to be $ln3/\bar{d}_s$ for the FS method. For RGB-only images, we increase $w_{7}$ by 6 times to compensate the increase of probabilities of selecting the FS and the SS method. The initial weights $\alpha_1$ and $\alpha_2$ in (\ref{eqn:elastic_net}) are set to be 1 and 2, respectively. The threshold $\omega_{min}$ and the step size $\tau$ in Algorithm \ref{alg:optimization} are set to be $1/3$ and $0.2$, respectively. We found that our model is insensitive to these parameters.

\subsection{Evaluation of the components of our method}
\label{sec:components}

\begin{figure}
  \centering
  \includegraphics[width=1.0\linewidth]{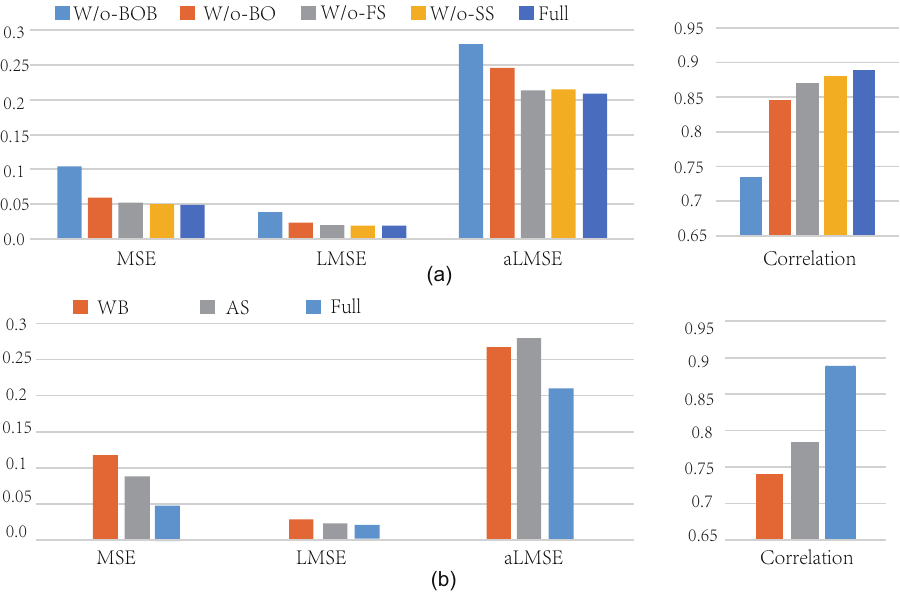}\\
  \captionsetup{font={footnotesize}}
  \caption{The effects of different components of our model on the MIT Intrinsic Images dataset. Larger values are better for the correlation, while smaller values are better for the other metrics. (a) The contributions of individual methods for estimating pairwise shading orders. (b) The roles of the brightening direction and the weights of pairwise shading orders.}\label{fig:Eva_components}
\end{figure}

\begin{figure}
  \centering
  \includegraphics[width=1.0\linewidth]{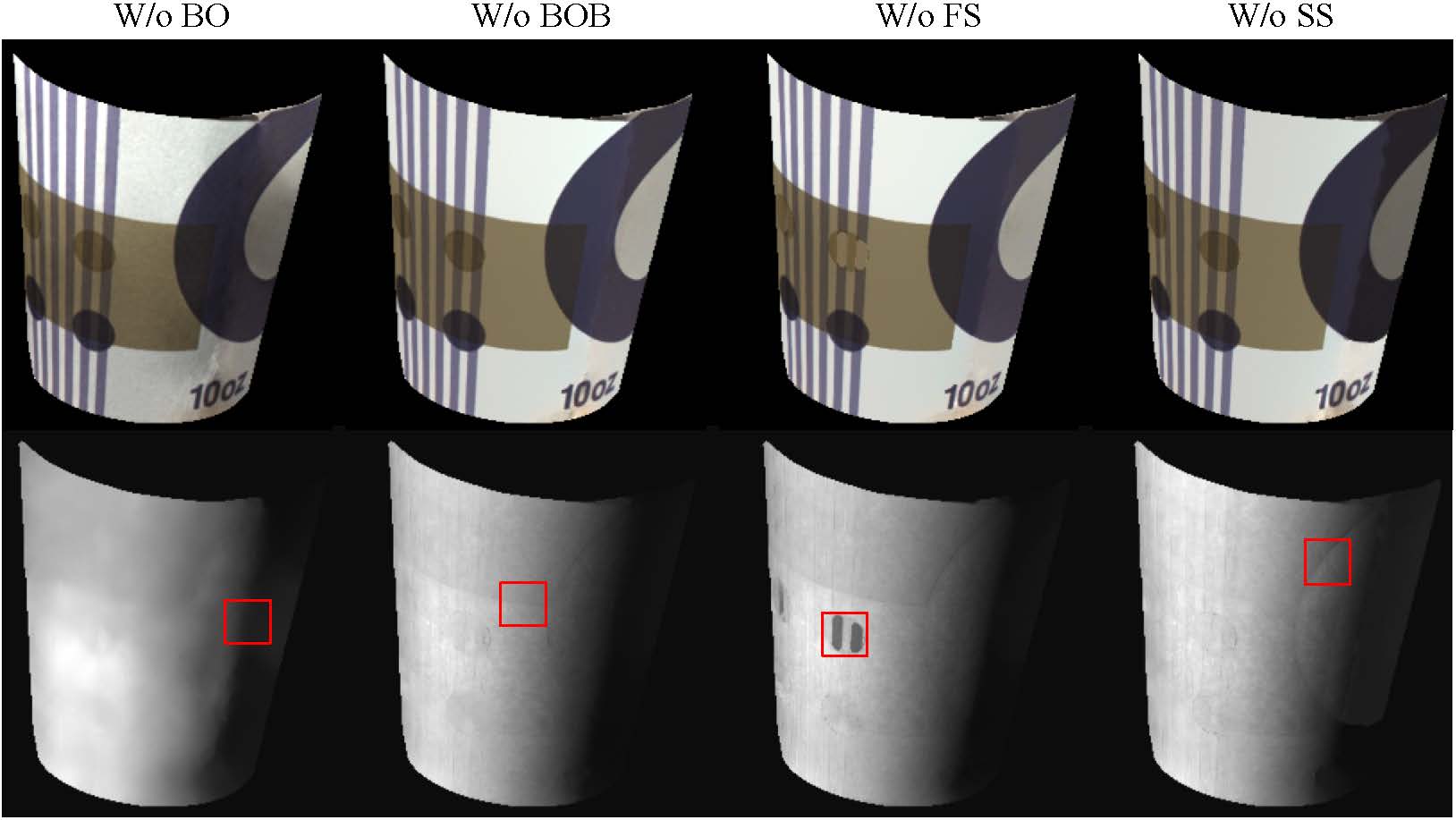}\\
  \captionsetup{font={footnotesize}}
  \caption{The recovered reflectance and shading when individual methods are unused. The red boxes point out the problematic shadings typically obtained from different configurations. See the text for more details.}\label{fig:Est_methods}
\end{figure}

\textbf{Individual estimation methods}. The results on the MIT Intrinsic Images dataset are compared in Fig. \ref{fig:Eva_components}a. Our full model (Full) achieves the best performance, while estimating the shading orders without any single method will cause a noticeable drop of performance. Disabling BOB (W/o BOB) causes the most severe drop, followed by BO, FS, and SS, consecutively. Fig. \ref{fig:Est_methods} shows the changes of the recovered reflectance and shading when different methods are removed. Removing BO will break the smoothness of reflectance across the shadow edges. When BOB is unused, the shading smoothness across different reflectance will be broken, leaving sharp edges in shading. The smoothness-based methods FS and SS are essential for keeping the local shading smooth. Without using FS, the smoothness in textured regions cannot be guaranteed. SS is important for the areas where the biases of reflectance brightness are not accurately estimated.

\noindent\textbf{The brightening direction}. We test a special case of our method, where the brightening direction is fixed at $[1, 1, 1]^T$ following the Color Retinex \cite{intrinsic_dataset}. Although the direct illuminants in the MIT Intrinsic Images dataset are nearly white and the ambient illuminants are weak, the performance under a white brightening direction (WB) is much worse than our original model (Fig. \ref{fig:Eva_components}b).

\noindent\textbf{The confidences of pairwise orders}. We evaluate the importance of the confidences of the pairwise orders in inferring the global shading by replacing AE with AS \cite{AngularSync}, \ie assigning equal weights to the pairwise shading orders. From Fig. \ref{fig:Eva_components}b we can see that the performance drops significantly.



\begin{figure}
  \centering
  \includegraphics[width=1.0\linewidth]{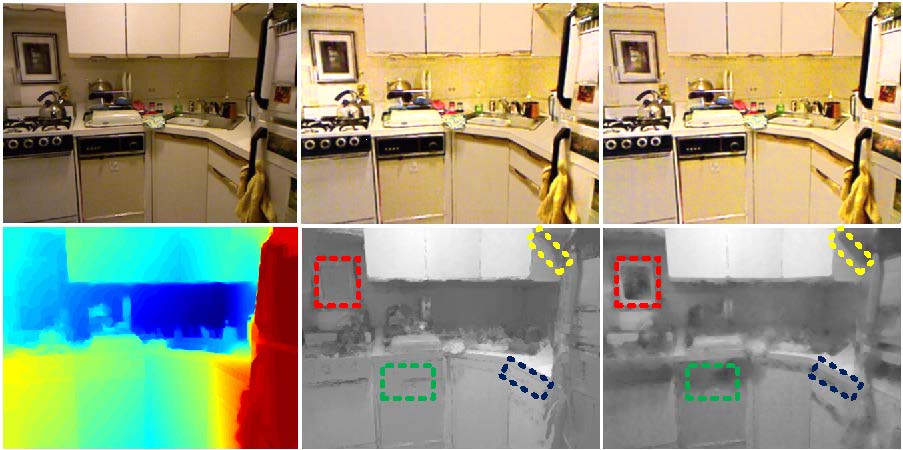}\\
  \captionsetup{font={footnotesize}}
  \caption{The effect of depth information. \textbf{Left}: The input RGB image and its depth map; \textbf{Middle}: The results when the depth map is taken as input; \textbf{Right}: The results without using depth information. The dashed boxes indicate some typical differences between the recovered shading.}\label{fig:depth}
\end{figure}

\noindent\textbf{Depth information}. Several depth-based features are used to calculate the confidences of pairwise orders for RGB-Depth images (Section \ref{sec:reliability}). Fig. \ref{fig:depth} suggests their effects. Utilizing the feature of Surface Normal Change increases the probability of applying the shading smoothness constraints to flat surfaces. See the regions in the red and green boxes of Fig. \ref{fig:depth} for examples. These areas are mistaken to be shadowed without depth cues, since they have similar chromaticity to their surroundings, and their boundaries are blurred. The feature of Shadow Edges finds shading changes at depth discontinuities efficiently. It may miss some shadow edges that cannot be generated by any sample of illuminant, when the change of depth is small (\eg the area in the blue box of Fig. \ref{fig:depth}), or a large part of the occluder is not visible in the current view (\eg the area in the yellow box).

\subsection{Results on MIT Intrinsic Images dataset}

\begin{table}
\captionsetup{font={footnotesize}}
\caption{Results on the MIT Intrinsic Images dataset. Higher correlation, lower MSE, LMSE and aLMSE are better.}
\label{tab:results}
\centering
\begin{tabular}{l|c|c|c|c} \hline
                                                & Correlation      & MSE        & LMSE              & aLMSE \\ \hline
Grey Retinex \cite{Retinex}                     & 0.6494           & 0.1205          &    0.0329        & 0.3373 \\
Tappen \etal \cite{TappenCVPR06}                & - & - &0.0390 & -\\
Color Retinex \cite{intrinsic_dataset} & 0.7146           & 0.1108          &    0.0286        & 0.2541 \\
Jiang-A \cite{correlation}      & 0.6184 & 0.1533 &0.0421 &0.3988\\
Jiang-H \cite{correlation}      & 0.5829 & 0.1524 &0.0483 &0.3476\\
Jiang-HA \cite{correlation}      & 0.6109 & 0.1579 &0.0454 &0.3631\\
Shen-SR \cite{Shen_PAMI13} & 0.7259 & 0.1223 & 0.0240 &0.2454 \\
Shen-SRC \cite{Shen_PAMI13}        & -            & -   &0.0204      & -       \\
Zhao \etal \cite{Zhao_PAMI13}        & -            & -   &0.0250      & -       \\
Gehler \etal \cite{gehler11nips}        & 0.7748            & 0.0985   &  0.0244      & 0.2544       \\
Serra \etal \cite{Serra_CVPR12}     & 0.7862            & 0.0834    &  0.0340      & 0.2958      \\
Bell \etal \cite{Bell_Siggraph14}   & 0.7229 & 0.1100 &0.0337 & 0.2763\\
Li \etal \cite{Li_2014_CVPR}                & - & - & 0.0190 & -\\
Chang \etal \cite{chang14svdpgmm}           & - & - &0.0229 & -\\
SIRFS \cite{BarronECCV12} & 0.8095 &0.0567 & 0.0279 & 0.2329\\
Ours-AE \cite{liu_accv2014}            & 0.8582 & 0.0684 &0.0189 & 0.2252 \\
Ours-CSF                        & \textbf{0.8867} & \textbf{0.0492} & \textbf{0.0186} & \textbf{0.2089} \\

\hline
\end{tabular}
\end{table}

\begin{table}[t!]
\captionsetup{font={footnotesize}}
\caption{Results on the IIW dataset evaluated by $WHDR_{10\%}$.}
\label{tab:results_IIW}
\centering
\setlength{\tabcolsep}{2pt}
\begin{tabular}{c*{6}{|c}} \hline
Retinex \cite{intrinsic_dataset} & Shen \cite{Shen_CVPR11}      & Garces \cite{Garces2012}    &  Zhao \cite{Zhao_PAMI13}  & Bell \cite{Bell_Siggraph14}              & Bi \cite{Bi2015L1Intrinsic} & Ours \\ \hline
27.4\% & 32.4\% & 25.9\% & 23.7\% & 21.1\% & 18.1\% & 19.8\% \\\hline
\end{tabular}
\end{table}

We compare our method to the state-of-art and to several classic approaches as listed in Table \ref{tab:results}. These results are either copied from their papers, the report in \cite{Serra_CVPR12}, or by running their code directly without tuning any parameters \footnote{The method SIRFS is evaluated on the images of cup2, deer, frog2, paper2, raccoon, sun, teabag1 and turtle, while the other images are used for training. The results of Bell \etal \cite{Bell_Siggraph14} are obtained through relaxing the constraints on the absolute values of shading and removing the intensity from the features for clustering the reflectance. Otherwise the deep shadows will be mistaken to be black and clustered into individual categories.}. We report the results under the best parameters for the whole dataset. Our method achieves the best performance.

\begin{figure*}
  \centering
  \includegraphics[width=1.0\linewidth]{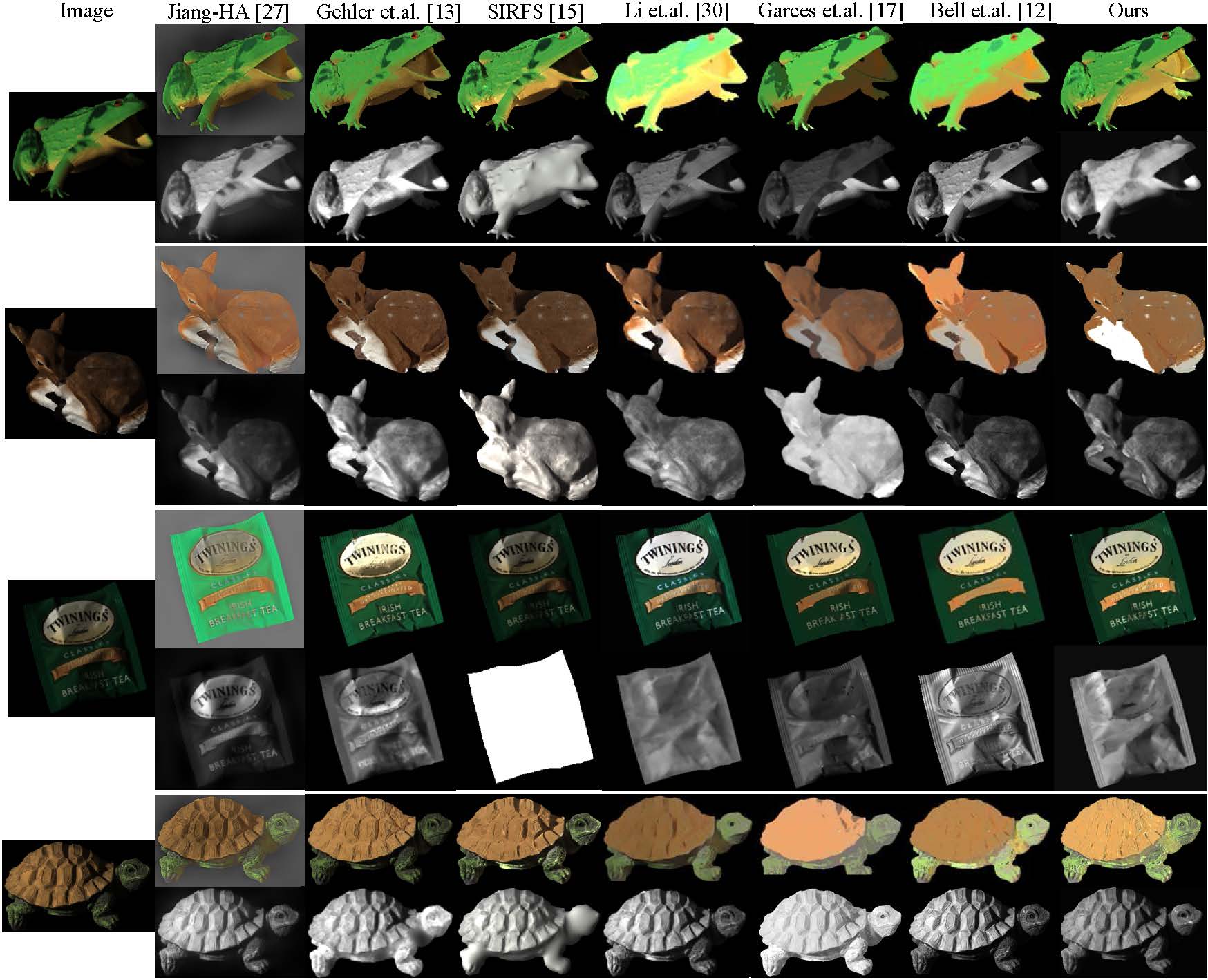}\\
  \captionsetup{font={footnotesize}}
  \caption{Typical results on the MIT Intrinsic Images dataset.}\label{fig:MIT}
\end{figure*}

Fig. \ref{fig:MIT} gives some concrete examples. The most remarkable advantage of our method is that it can recover the reflectance under deep shadows. One reason is that we can cluster the pixels with the same reflectance together on the $UV$ shadow-free plane, no matter how dramatically the shading changes. Another reason is that our model fuses estimates from different methods by selecting the optimal one exclusively, which avoids smoothing the shading edges by the other estimates. Clustering-based methods, including Gehler \etal \cite{gehler11nips}, Garces \etal \cite{Garces2012}, and Bell \etal \cite{Bell_Siggraph14}, are sensitive to the change of intensity and color caused by shadows. The edge-based method of Li \etal \cite{Li_2014_CVPR} tends to assign large gradients to reflectance changes, which degrades at sharp shadow edges (\eg those on the body of the deer). The methods of Gehler \etal \cite{gehler11nips} and Li \etal \cite{Li_2014_CVPR} smooth the shading extensively, leaving residuals of shadows in the reflectance (\eg the teabag). SIRFS \cite{BarronECCV12} smoothes the surfaces, which may generate an overly smooth shading (\eg the frog).

Another advantage is that our method can recover the global shading robustly. The main reason is that the clustering-based methods BO and BOB capture the shading orders between distant pixels effectively. Edge-based methods cannot reliably recover the relative shading between unconnected parts (\eg the shadings recovered by Li \etal \cite{Li_2014_CVPR} are inconsistent between the front and the back of the turtle). Another reason is that BOB can handle the areas where the shading and reflectance change simultaneously (\eg the mouth and the head of the frog).

Our method preserves the subtle variations of reflectance (\eg the yellow and orange regions of the tea bag), since the intra-cluster variations in the $UV$ plane (Fig. \ref{fig:brighten}c) are represented in the recovered reflectance. In contrast, some clustering-based methods, such as Garces \etal \cite{Garces2012} and Bell \etal \cite{Bell_Siggraph14}, unify the reflectance of the pixels of each cluster. This operation often leads to block artifacts (\eg the tea bag).

Our method did not handle the feet of the deer well. The black feet and the white legs are both achromatic, so they fall into the same cluster on the shadow-free plane. The image blur further reduces the efficiency of the feature of Reflectance Change (Section \ref{sec:reliability}), so the difference between black and white are not kept into reflectance.

\subsection{Results on Natural Images}
\label{sec:IIW}

The quantitative results on the IIW dataset are shown in Table \ref{tab:results_IIW}. Our method achieved comparable results to the state-of-art. It should be mentioned that $WHDR_{10\%}$ cannot reflect the superiority of our method on inferring the shading orders between pixels with different chromaticity, since only pixels with similar chromaticity are compared \cite{Bell_Siggraph14}. Further, the textured pixels are excluded from evaluation, so the ability to preserve the texture of reflectance is untested. Actually, both the top-performing methods of \cite{Bi2015L1Intrinsic} and \cite{Bell_Siggraph14} remove the texture from the reflectance. For a fair comparison, we report our result that uses the edge-preserving smoothing of \cite{Bi2015L1Intrinsic} to preprocess the input image. Without smoothing, the $WHDR_{10\%}$ increases about $3.7\%$.

\begin{figure*}
  \centering
  \includegraphics[width=1.0\linewidth]{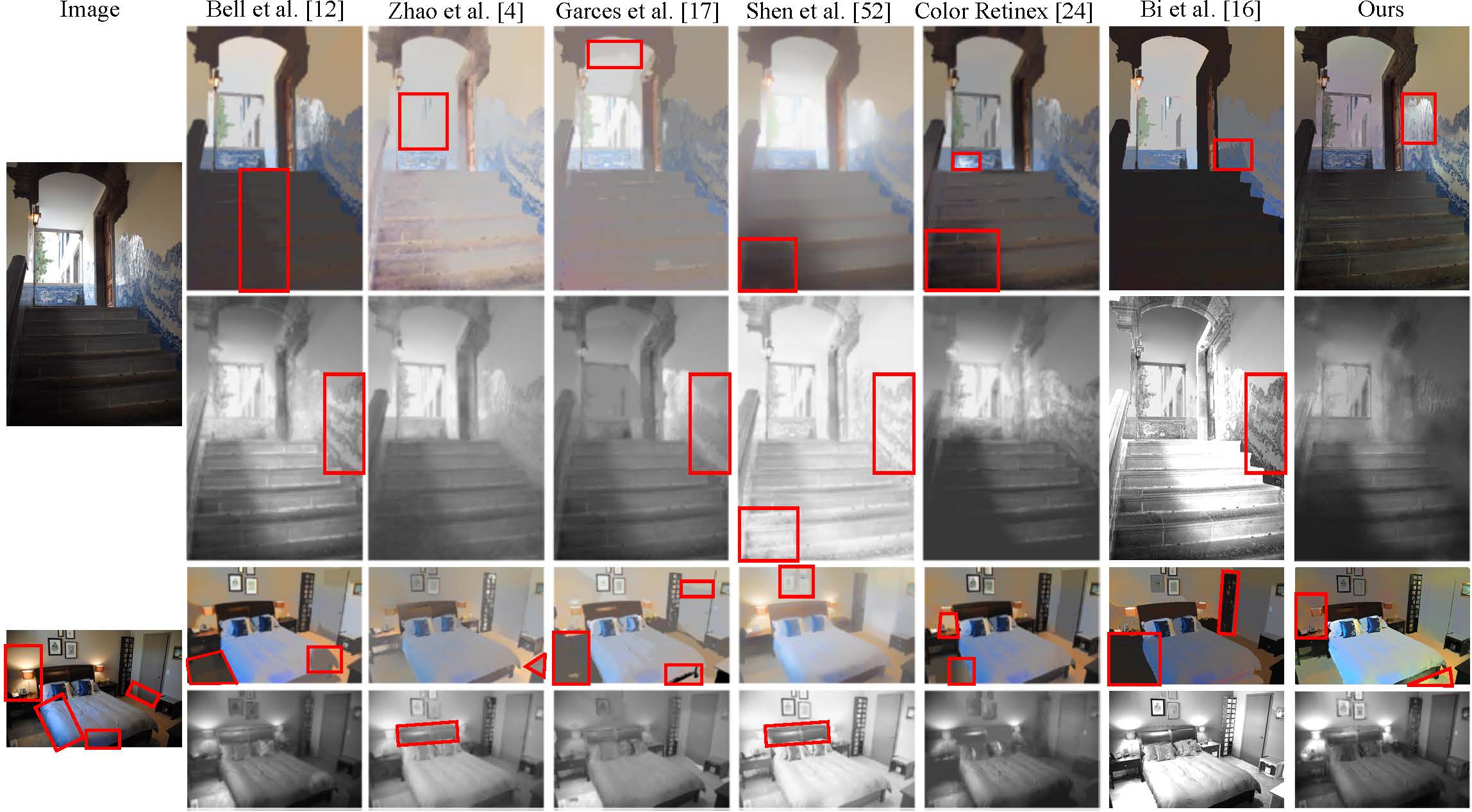}\\
  \captionsetup{font={footnotesize}}
  \caption{Representative results for the IIW dataset. More examples are given in the supplementary.}\label{fig:results_IIW}
\end{figure*}

The IIW dataset is much more difficult than the MIT Intrinsic Images dataset. The image in the top row of Fig. \ref{fig:results_IIW} is comprised of different kinds of objects, some of which are highly textured (\eg the wall with blue painting). Our method preserves the textures \footnote{We do not use the edge-preserving smoothing to produce the qualitative results in Fig. \ref{fig:results_IIW}.} much better than the other methods in comparison. Another difficulty comes from the intensive specular reflections (\eg the wall in the top row of Fig. \ref{fig:results_IIW}). Our method puts the specular reflections into reflectance, while some other methods, such as Zhao \etal \cite{Zhao_PAMI13} and Garces \etal \cite{Garces2012}, put them into shading.

The greatest challenge of the IIW dataset comes from the coexistence of multiple direct illuminants in the same scene. In the bottom row of Fig. \ref{fig:results_IIW}, the areas in the red boxes of the input image are covered by lights in different colors. This situation does not satisfy the bi-illuminant assumption of the BIDR model \cite{BIDR}. No unique brightening direction exists for the whole image, and the brightening direction obtained from entropy minimization (Section \ref{sec:brightness}) eliminates the difference improperly. It causes two problems to our method: (1) the error of clustering will increase; and (2) the color of the recovered reflectance will be twisted. The first problem is shared by all the clustering-based methods such as Garces \etal \cite{Garces2012} and Bell \etal \cite{Bell_Siggraph14}. The second problem is common, since all the methods in comparison assume a single (direct) illumination. Despite these problems, our model still recovered a globally consistent shading.

\textbf{Discussion}. Scene-SIRFS addressed the mixture of illuminations by a soft segmentation of the image with respect to the "ownership" of illuminants \cite{Barron2013A}. But the segmentation is not easy, since the changes of illuminations are often slower than the changes of reflectance. Beigpour and Van de Weijer \cite{Beigpour_ICCV11} proposed the Multi-illuminant Dichromatic Reflection (MIDR) model to account for the secondary illuminants. However, in practice they only dealt with the case of two direct illuminants irradiating a single-colored object. We may consider extending the BIDR model to incorporate multiple direct illuminants. Accordingly, there will be multiple brightening directions, and the brightness should be extended to a mixture of sub-coordinates. This will make the problem much more complex.

\begin{figure}
  \centering
  \includegraphics[width=1.0\linewidth]{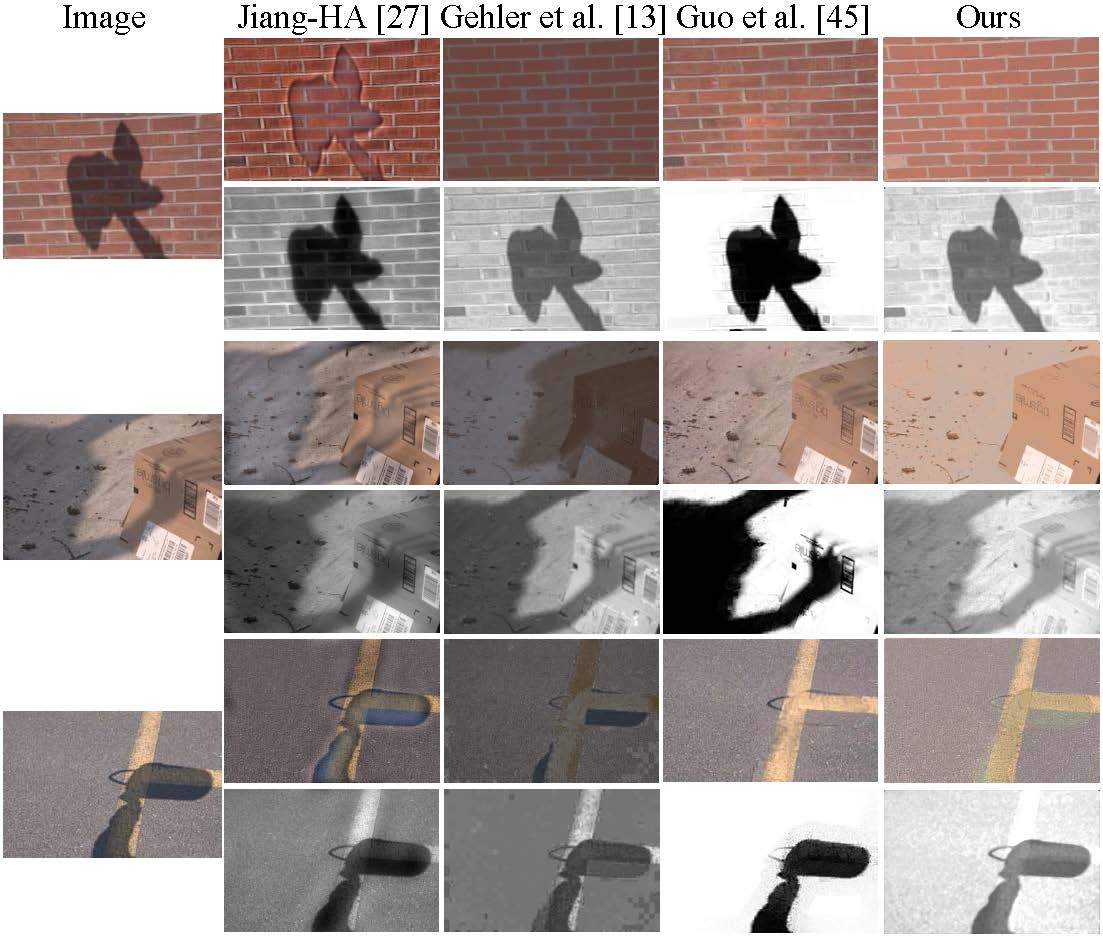}\\
  \captionsetup{font={footnotesize}}
  \caption{Typical results on the UIUC shadow dataset. More examples are shown in the supplementary.}\label{fig:shadow_removal}
\end{figure}

We further test on the outdoor images from the UIUC shadow dataset \cite{Guo_PAMI12}. Fig. \ref{fig:shadow_removal} shows three examples. The ambient illuminants are usually the blue sky, so the shadowed areas are more blueish than the lit areas. We compare to the methods of Jiang-HA \cite{correlation} and Gehler \etal \cite{gehler11nips}. We also compare to the region-pair-based shadow removal method proposed by Guo \etal \cite{Guo_PAMI12}\footnote{For this method the shading is replaced by a shadow map, in which black pixels indicate shadows and gray ones stand for penumbra.}. Our model recovers the reflectance by lighting the dark pixels along the yellowish brightening direction, while the other intrinsic decomposition methods often fail to recover their colors. The method of Guo \etal \cite{Guo_PAMI12} is unable to handle thin areas due to the limited resolution of image segmentation (\eg the fingers in the last image of Fig. \ref{fig:shadow_removal}).


\subsection{Evaluation on RGB-Depth Images}
\label{sec:RGB_D}

\begin{figure*}
  \centering
  \includegraphics[width=1.0\linewidth]{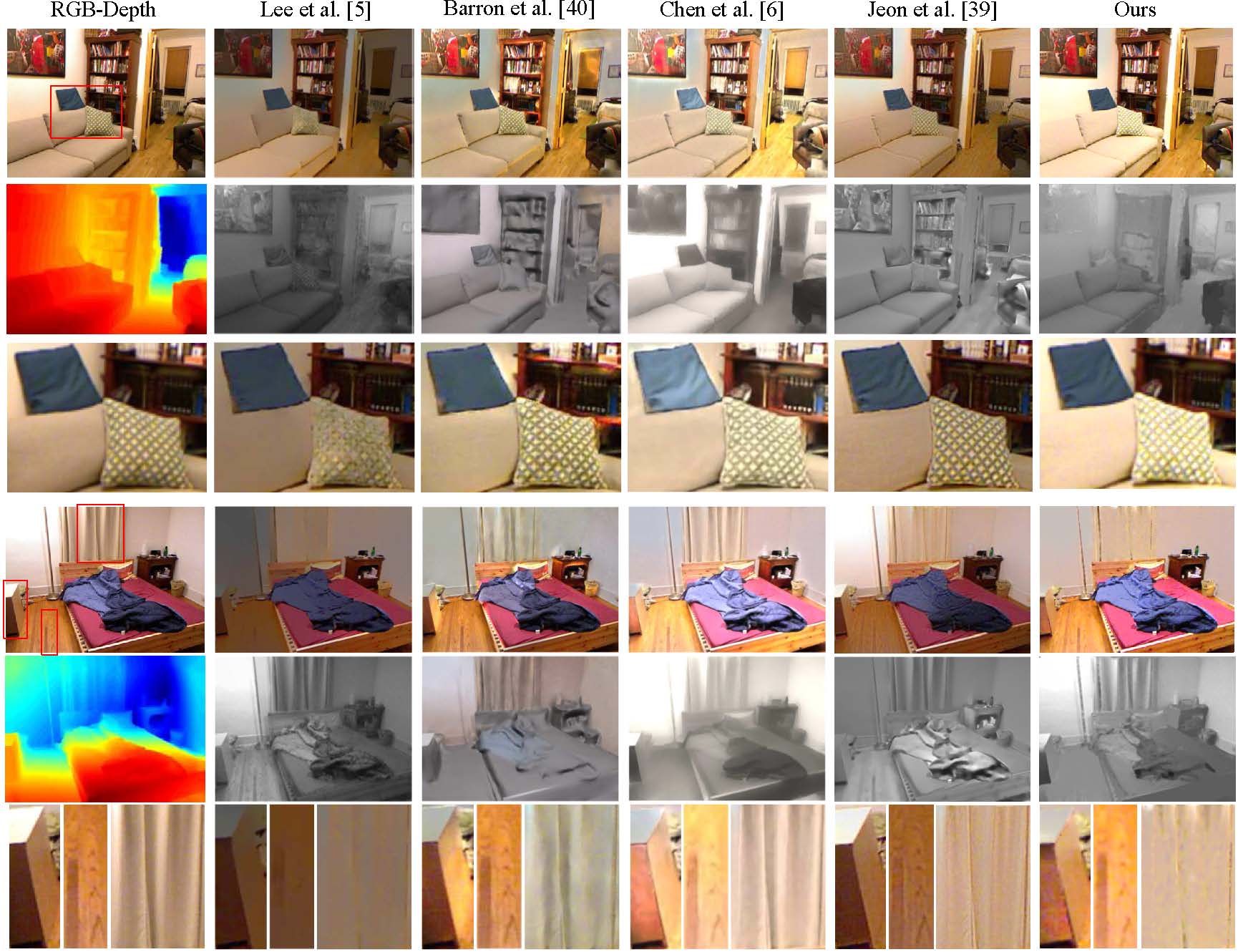}\\
  \captionsetup{font={footnotesize}}
  \caption{Representative results for the NYU-Depth V2 dataset. More examples are given in the supplementary.}\label{fig:results_NYU}
\end{figure*}

We test on the RGB-Depth images from the NYU-Depth V2 dataset. We compare to those methods that take RGB-Depth images  \cite{Barron2013A}\cite{Chen_ICCV13}\cite{Jeon14texture} or videos \cite{Lee_ECCV12} as input\footnote{We combine the direct irradiance, the indirect irradiance and the illumination into shading if needed. The temporal constraints of \cite{Lee_ECCV12} are removed for dealing with single images.}. Typical examples are shown in Fig. \ref{fig:results_NYU}. Our method successfully recovered globally consistent shadings and preserves the textures of reflectance. In particular, our method was the only one that recovers the smooth shading over the painting in the first row of Fig. \ref{fig:results_NYU}. In comparison, the method of Lee \etal \cite{Lee_ECCV12} did not get consistent shadings between surfaces in different orientations. In their recovered reflectance of the first image in Fig. \ref{fig:results_NYU}, the backrest of the sofa and the walls are much darker than the seat of the sofa and the floor.

The method of Barron and Malik \cite{Barron2013A} successfully captured the shape of curved surfaces (\eg the sofa in the first image of Fig. \ref{fig:results_NYU}) but not those of objects with sharp boundaries (\eg the cabinet and the bed in the second image of Fig. \ref{fig:results_NYU}). The method of Chen and Koltun \cite{Chen_ICCV13} achieved good smoothness of shading while keeping the sharp surface edges at the same time. However, this method often failed to recover the shading orders between objects with different colors (\eg the blue pillow and the sofa in the first image of Fig. \ref{fig:results_NYU}). The method of Jeon \etal \cite{Jeon14texture} preserved the textures in reflectance very well (\eg the floor in the second image of Fig. \ref{fig:results_NYU}), but this method tends to reduce the difference of shading between surfaces with similar orientations (\eg the walls and the cabinet in the second image of Fig. \ref{fig:results_NYU}).

\section{Conclusions and discussions}

We proposed the shading orders for intrinsic image decomposition. The shading orders captured not only adjacent relations but also distant connections. This overcame the limitations of edge-based methods that lack the large-scale structure of shading. The shading orders can be measured by several individual methods, each of which can give a reasonable estimate based on certain assumptions about the scene. Jointly utilizing these methods captured various kinds of priors and observations of the scene.

We developed the CSF algorithm to combine the pairwise orders measured by different methods. CSF infers a global order by selecting the confident and consistent pairwise orders and solving their conflicts through AE. The local competition removes unreliable measurements from the fusion, so the results are much cleaner than a weighted sum of different estimates. This is essential for keeping sharp shadow edges and textures. The sparsity-driven neighbor selection further reduced the outliers of local measurements.

Experimental results demonstrated that our model is suitable for various indoor and outdoor scenes with noticeable ambient illuminants. However, the BIDR model cannot handle multiple direct illuminants, interreflections, or specular reflections. We need to generalize the BIDR model and the $UVB$ color space for more realistic scenes.

The highly textured images are still quite challenging for clustering-based methods, since their reflectance often changes irregularly and thus cannot be clustered properly. Jeon \etal proposed to separate the texture layer before decomposing the shading and reflectance \cite{Jeon14texture}, which is a promising way to ease the clustering.

\appendix
\section*{Rendering the shading map}
\begin{figure*}
  \centering
  \includegraphics[width=1.0\linewidth]{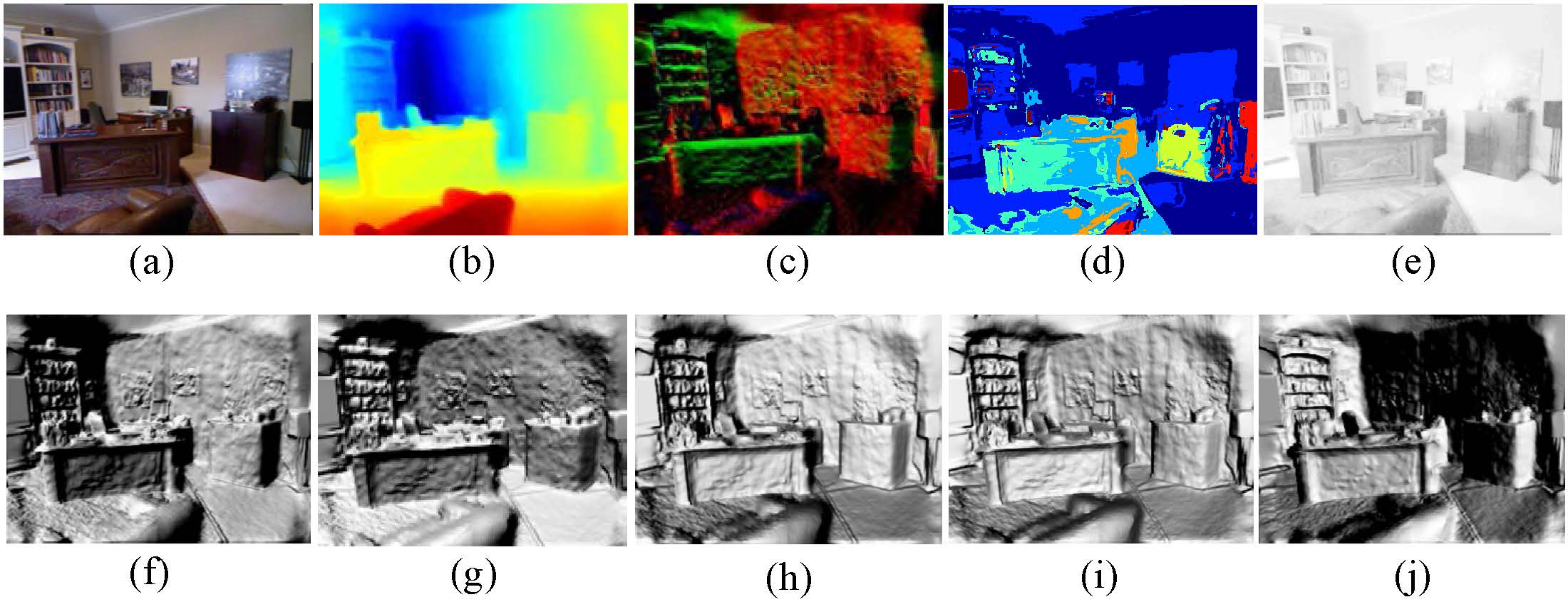}\\
  \caption{The geometry of the scene and its shading rendered under illuminants at different positions. (a) The raw image; (b) The depth map; (c) The surface normal map derived from the depth map; (d) The cluster map based on chromaticity; (e) The image brightness; (f)-(j) The shading rendered under direct illuminants at positions of (110, -176, -251), (510, -176, 49), (-389, -176, -151), (-191, 176, -751) and  (510, -476, -951), respectively. Their similarities to the image brightness are 0.27, 0.20, 0.11, 0.00, and -0.26, respectively. The positions of the illuminants are specified by $(x, y, z)$ in the camera coordinate, where the $x$ and $y$ axes are aligned with the horizontal and vertical directions of the image, respectively. The positive direction of the $x$ axis is from right to left. The positive direction of the $y$ axis is from top to bottom. The positive direction of the $z$ axis is from the camera to the image plane. The camera's position is set to be (0, 0, 0).}
  \label{fig:light_position}
\end{figure*}

Fig. \ref{fig:light_position} shows the rendered shading map of an RGB-Depth image. In the camera coordinate, we draw a "gray surface", taking all the pixels as vertices. Both the color of the surface and the illuminant are set to be $[1, 1, 1]^T$, and the reflection of the surface is set to be diffuse only (\ie without any specular reflection). Here we assume that there is only one direct illuminant for each image, while the ambient illumination is set to be 0. The illuminant is put inside the room box, and the range of the room box is set to be the scope of all the observable pixels. Especially, we expand the range of the $z$ dimension (orthogonal to the image plane) to the negative part of the coordinate, since the light may be placed to the back of the camera. The surface is rendered with the Matlab Surfl function, and the output intensities of the vertices form a shading map. The bottom row of Fig. \ref{fig:light_position} shows the rendering results under several sampled illuminants. We can see that some of them are close to the real shading map of the scene, while the others are quite different.

The similarity between the rendered shading and the ground-truth shading brightness $S^b$ is measured by their category-wise correlation:
\begin{equation}
\small
\label{eqn:Sim}
\begin{aligned}
Sim(\tilde{\gamma}(\L_d), S^b) &= \sum_{g\in G} \frac{n_g}{n} Cor(\tilde{\gamma}_g(\L_d), e^{S^b_g})\\
                               &= \sum_{g\in G} \frac{n_g}{n} Cor(\tilde{\gamma}_g(\L_d), e^{I^b_g}),
\end{aligned}
\end{equation}
where $G$ is the set of reflectance categories, $n$ is the number of pixels, and $Cor$ is the correlation between two variables. The subscripts $g$ denotes the subset of pixels belonging to the $g$-th category. Here we utilized the linear relationship between the brightness $I^b$ and the shading brightness $S^b$ based on (6). We select a set of candidate illuminants $\mathcal{L}=\{\L_d|Sim(\tilde{\gamma}(\L_d), S^b)>0.2\}$.

\section*{ADMM for optimizing the weights $W$}
Eqn. \ref{eqn:AE_matrix} can be solved for each pixel $p$ individually, where the matrix $W$ can be decomposed into a series of vectors $W_{p,\cdot}$. So do $E$ and $\tilde{C}$. For simplicity, we omit the subscript $p$ of all the matrixes from now on. Denote $d=D_p$. We reformulate Eqn. \ref{eqn:AE_matrix} to an equivalent problem:
\begin{equation}
\label{eqn:optimization_W}
\begin{aligned}
&\arg\min_{W,X,Y} g_1(W)+g_2(X)+g_3(Y)\\
&\mbox{s.t. } \tilde{C}^T W = d\\
&\ \ \ \ \ \   W=X=Y
\end{aligned}
\end{equation}
where
\begin{equation}
\label{eqn:g}
\begin{aligned}
g_1(W)&=E^T W + \frac{\alpha_2}{2}\|W\|_2^2\\
g_2(X)&=\alpha_1 \|X\|_1\\
g_3(Y)&=\left\{
\begin{aligned}
&0  &\mbox{if } Y_q \geq 0, \forall q \\
&\infty &\mbox{otherwise}\\
\end{aligned}\right.
\end{aligned}
\end{equation}

Through introducing Lagrange multipliers $\lambda$, $\Gamma_1$ and $\Gamma_2$, we can obtain the following augmented Lagrangian \cite{ADMM}:
\begin{equation}
\begin{aligned}
\label{eqn:L_W}
\mathcal{L}(W,X,Y,\lambda,\Gamma_1,\Gamma_2)=& g_1(W)+g_2(X)+g_3(Y)\\
&+ \lambda (d - \tilde{C}^T W) \\
&+ \Gamma_1^T (W-X)+\frac{\rho}{2}\|W-X\|_2^2\\
&+\Gamma_2^T (W-Y)+\frac{\rho}{2}\|W-Y\|_2^2
\end{aligned}
\end{equation}
where $\rho$ is a scaling parameter. We initialize $W$, $X$ and $Y$ with $1_n$, while $\lambda=2$ and $\Gamma_1 = \Gamma_2=1_n$.
Then we update them iteratively as follows:
\begin{equation}
\label{eqn:update_vars}
\begin{aligned}
W^{k+1}&=\frac{1}{\alpha_2+2\rho}(\rho X^{k}+\rho Y^{k}-E+\lambda^{k}\tilde{C}-\Gamma_1^{k}-\Gamma_2^{k})\\
X^{k+1}&=\left\{
\begin{aligned}
&W^{k+1}+\Gamma_1^{k}-\frac{1}{\rho}\alpha_1  &\mbox{if } W^{k+1}+\Gamma_1^{k}>\frac{1}{\rho}\alpha_1 \\
&0  &\mbox{if } |W^{k+1}+\Gamma_1^{k}|\leq\frac{1}{\rho}\alpha_1 \\
&W^{k+1}+\Gamma_1^{k}+\frac{1}{\rho}\alpha_1  &\mbox{if } W^{k+1}+\Gamma_1^{k}<-\frac{1}{\rho}\alpha_1
\end{aligned}
\right.\\
Y^{k+1}&=(W^{k+1}+\frac{1}{\rho}\Gamma_2^{k})_+\\
\lambda^{k+1} &= \lambda^{k} + \eta_1 (d - \tilde{C}^T W^{k+1})\\
\Gamma_1^{k+1} &= \Gamma_1^{k} + \eta_2 (W^{k+1} - X^{k+1})\\
\Gamma_2^{k+1} &= \Gamma_2^{k} + \eta_3 (W^{k+1} - Y^{k+1})\\
\end{aligned}
\end{equation}
where $X$ is got from soft thresholding. $(\cdot)_+$ truncates all the elements of a vector to be non-negative. $\eta_1$, $\eta_2$ and $\eta_3$ are step sizes. We terminate the iteration when $\|W-X\|_1+\|W-Y\|_1$ is less than a threshold $T_W$ and $d - E^T W$ is less than a threshold $T_d$. In implementation, we set $\rho$, $\eta_1$, $\eta_2$ and $\eta_3$ to be 5, 0.05, 1, and 1, respectively.

\begin{figure}
  \centering
  \includegraphics[width=1.0\linewidth]{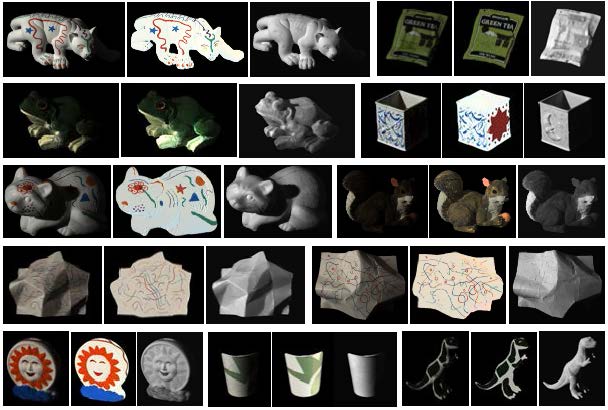}\\
  \caption{The results of our method for the rest 11 images of the MIT Intrinsic Images dataset. For each object, from \textbf{Left} to \textbf{Right}: The input image, the recovered reflectance, and the recovered shading.
  }\label{fig:MIT_suppl}
\end{figure}

\begin{figure}
  \centering
  \includegraphics[width=1.0\linewidth]{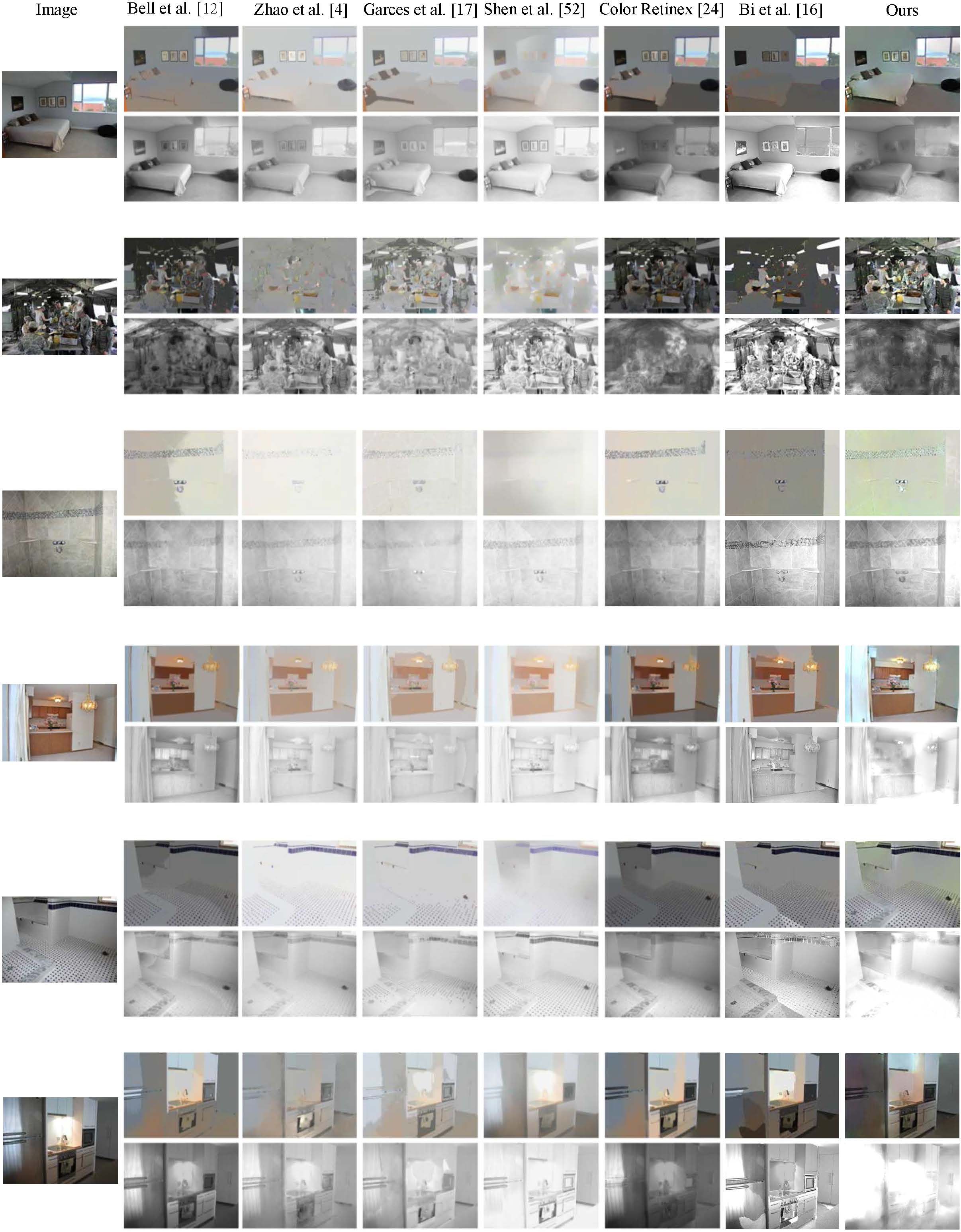}\\
  \caption{Representative results for the IIW dataset.
  }\label{fig:IIW_suppl_1}
\end{figure}

\begin{figure}
  \centering
  \includegraphics[width=1.0\linewidth]{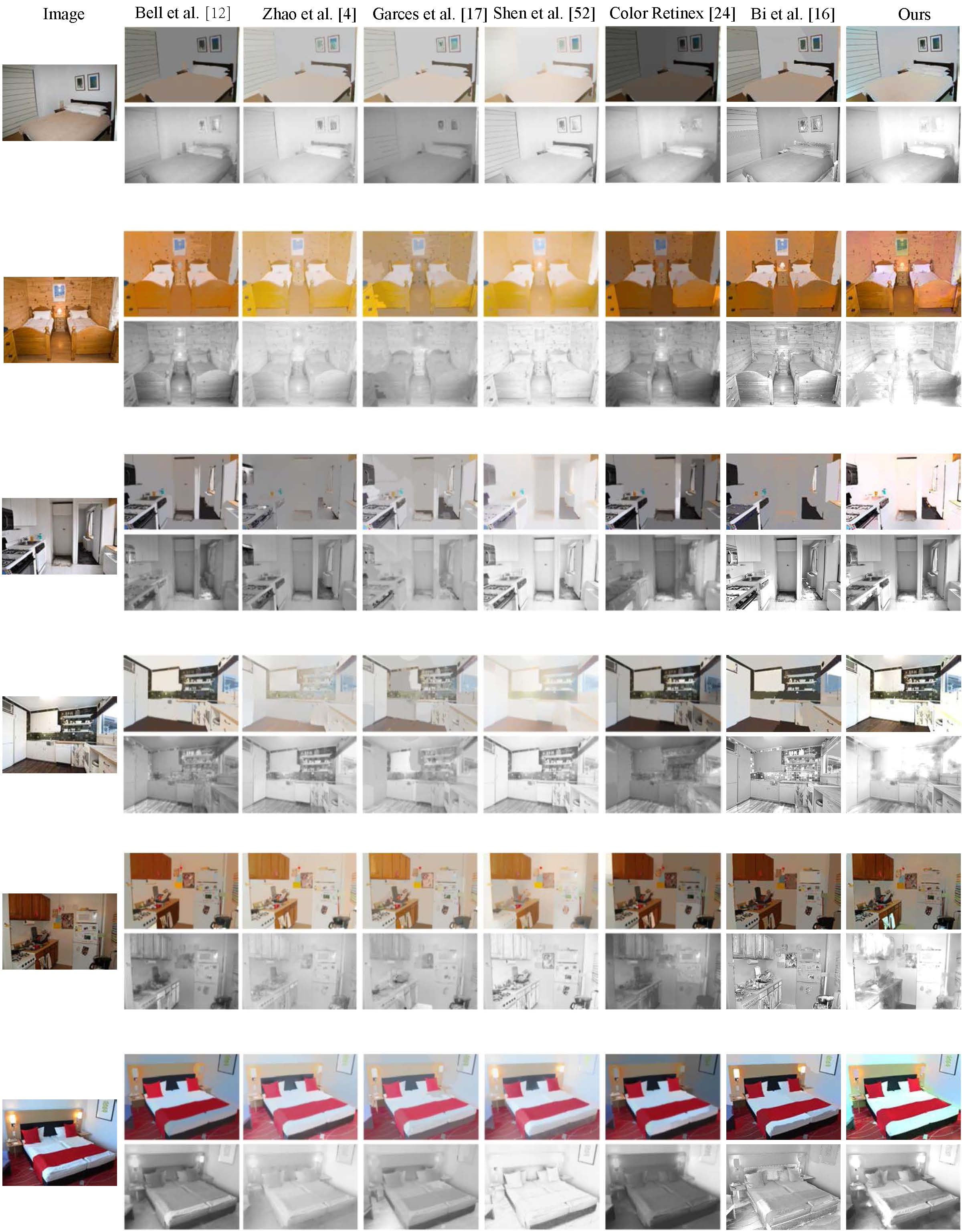}\\
  \caption{More representative results for the IIW dataset.
  }\label{fig:IIW_suppl_2}
\end{figure}

\begin{figure}
  \centering
  \includegraphics[width=1.0\linewidth]{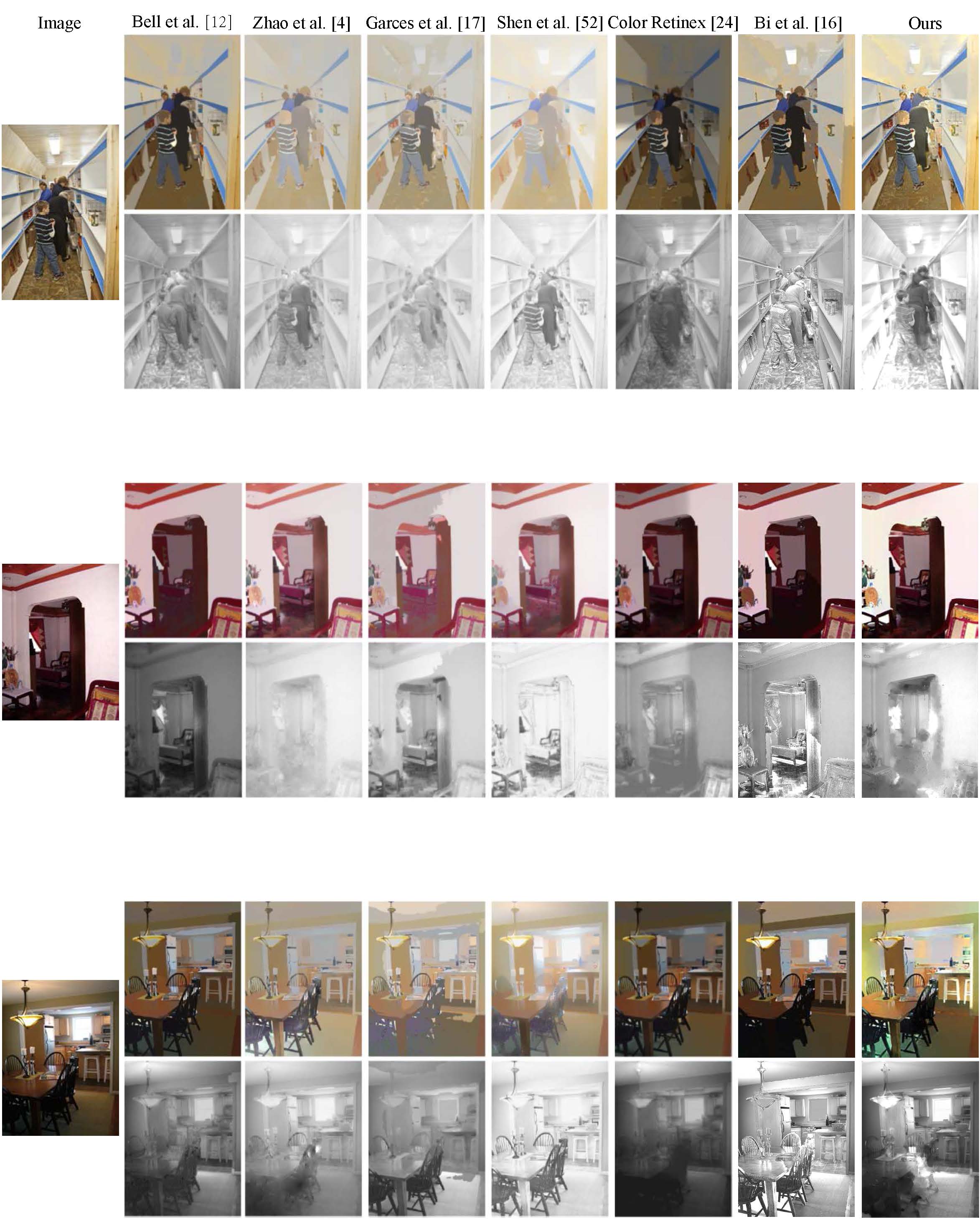}\\
  \caption{Even more representative results for the IIW dataset.
  }\label{fig:IIW_suppl_3}
\end{figure}

\begin{figure}
  \centering
  \includegraphics[width=1.0\linewidth]{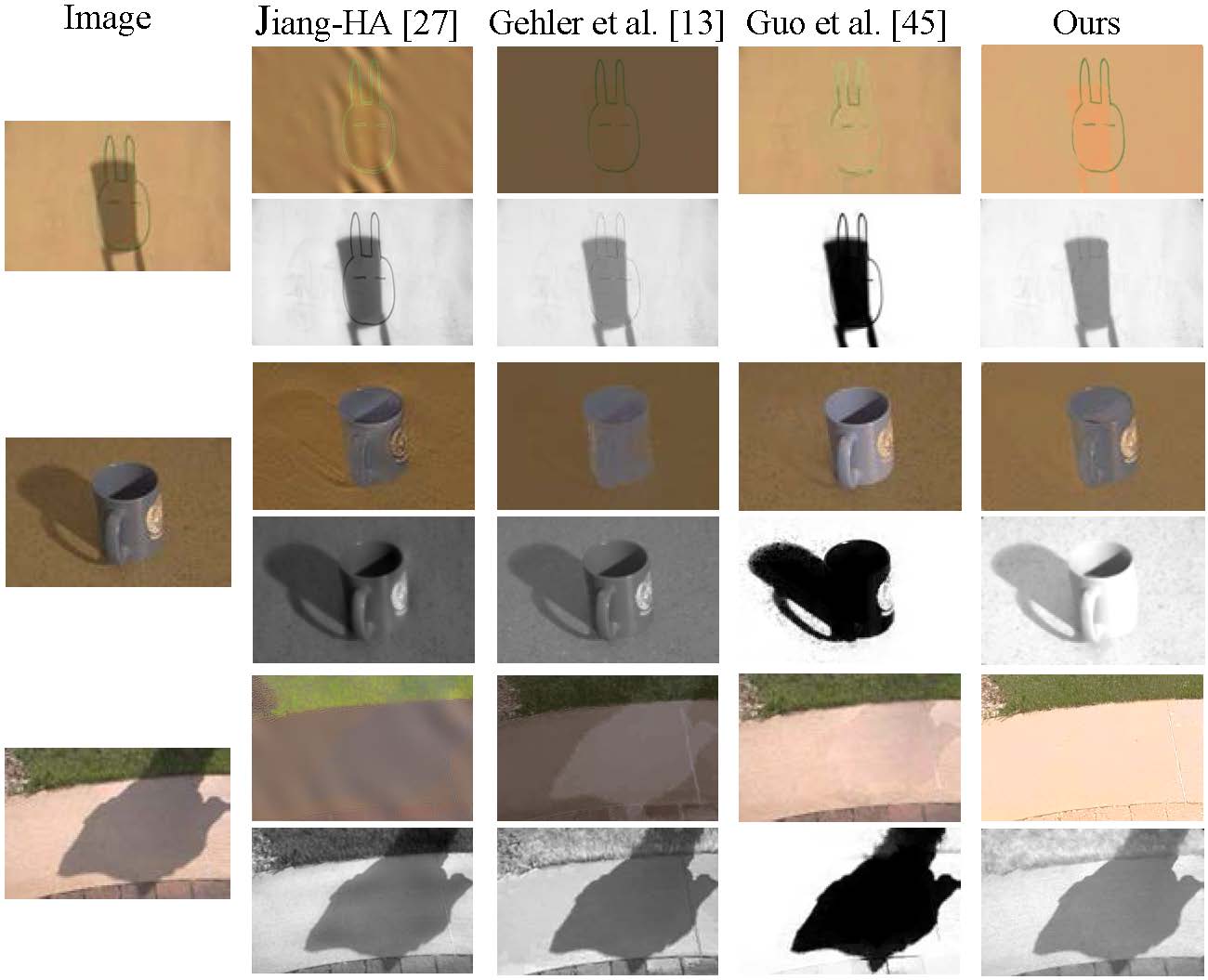}\\
  \caption{Typical results on the UIUC Shadow Removal dataset.
  }\label{fig:shadow_removal_suppl}
\end{figure}

\begin{figure}
  \centering
  \includegraphics[width=1.0\linewidth]{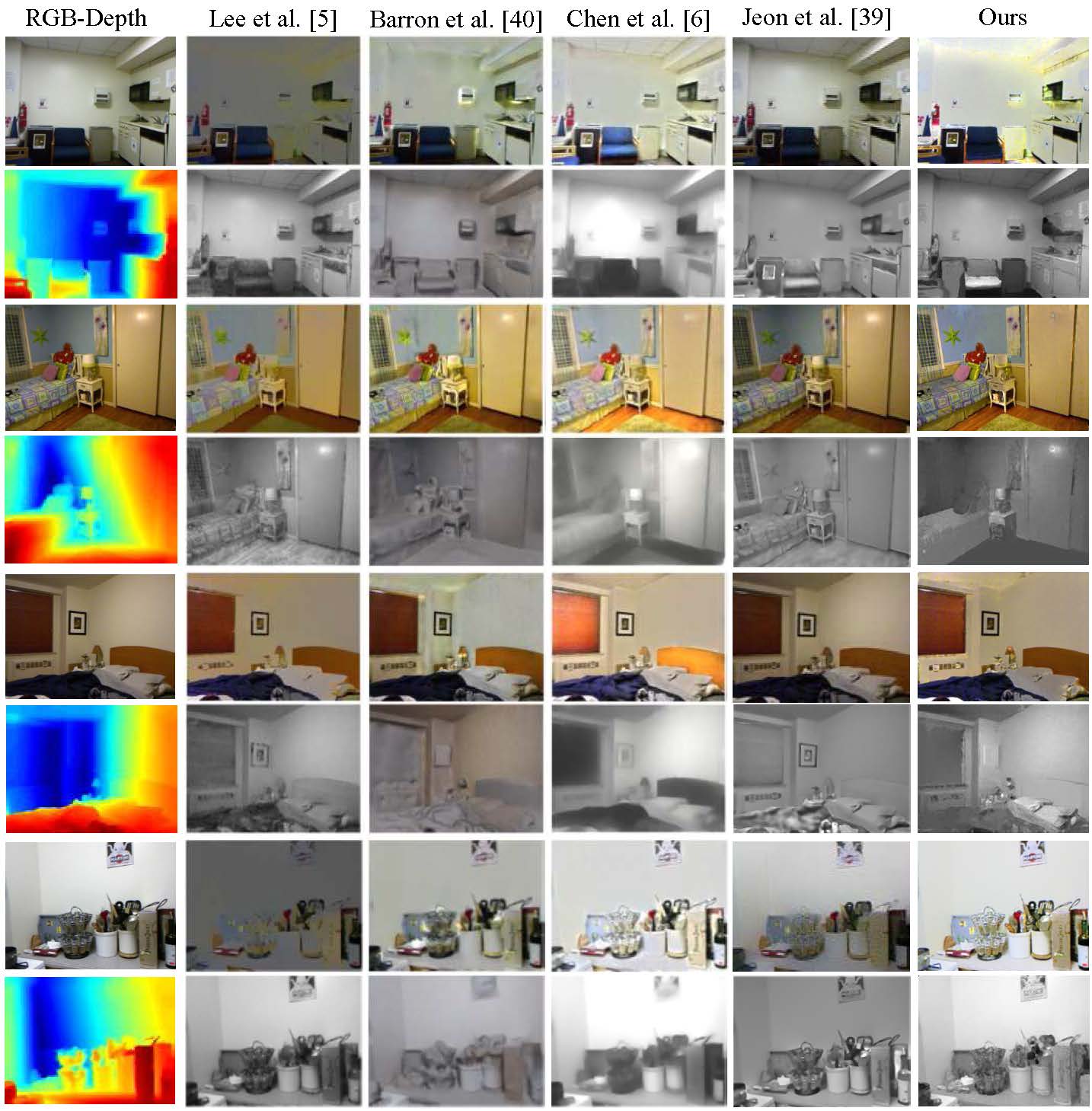}\\
  \caption{Representative results for the NYU-Depth V2 dataset.
  }\label{fig:NYU_suppl}
\end{figure}

Fig. \ref{fig:MIT_suppl} shows the results of our method for the images of the MIT Intrinsic Images dataset other than those appeared in the paper. Figs. \ref{fig:IIW_suppl_1}, \ref{fig:IIW_suppl_2} and \ref{fig:IIW_suppl_3} present several examples from the IIW dataset. Fig. \ref{fig:shadow_removal_suppl} shows more results of our method on the UIUC Shadow Removal dataset. Fig. \ref{fig:NYU_suppl} shows more results of our method on the NYU-Depth V2 dataset. We compare our method to several recent algorithms, including Bell \etal \cite{Bell_Siggraph14}, Zhao \etal \cite{Zhao_PAMI13}, Garces \etal \cite{Garces2012}, Shen \etal \cite{Shen_CVPR11}, Color Retinex \cite{intrinsic_dataset}, Bi \etal \cite{Bi2015L1Intrinsic}, Jiang-HA \cite{correlation}, Gehler \etal \cite{gehler11nips}, Guo \etal \cite{Guo_PAMI12}, Lee \etal \cite{Lee_ECCV12}, Barron \etal \cite{Barron2013A}, Chen \etal \cite{Chen_ICCV13}, and Jeon \etal \cite{Jeon14texture}.

\section*{Color of Shading}
\begin{figure}
  \centering
  \includegraphics[width=1.0\linewidth]{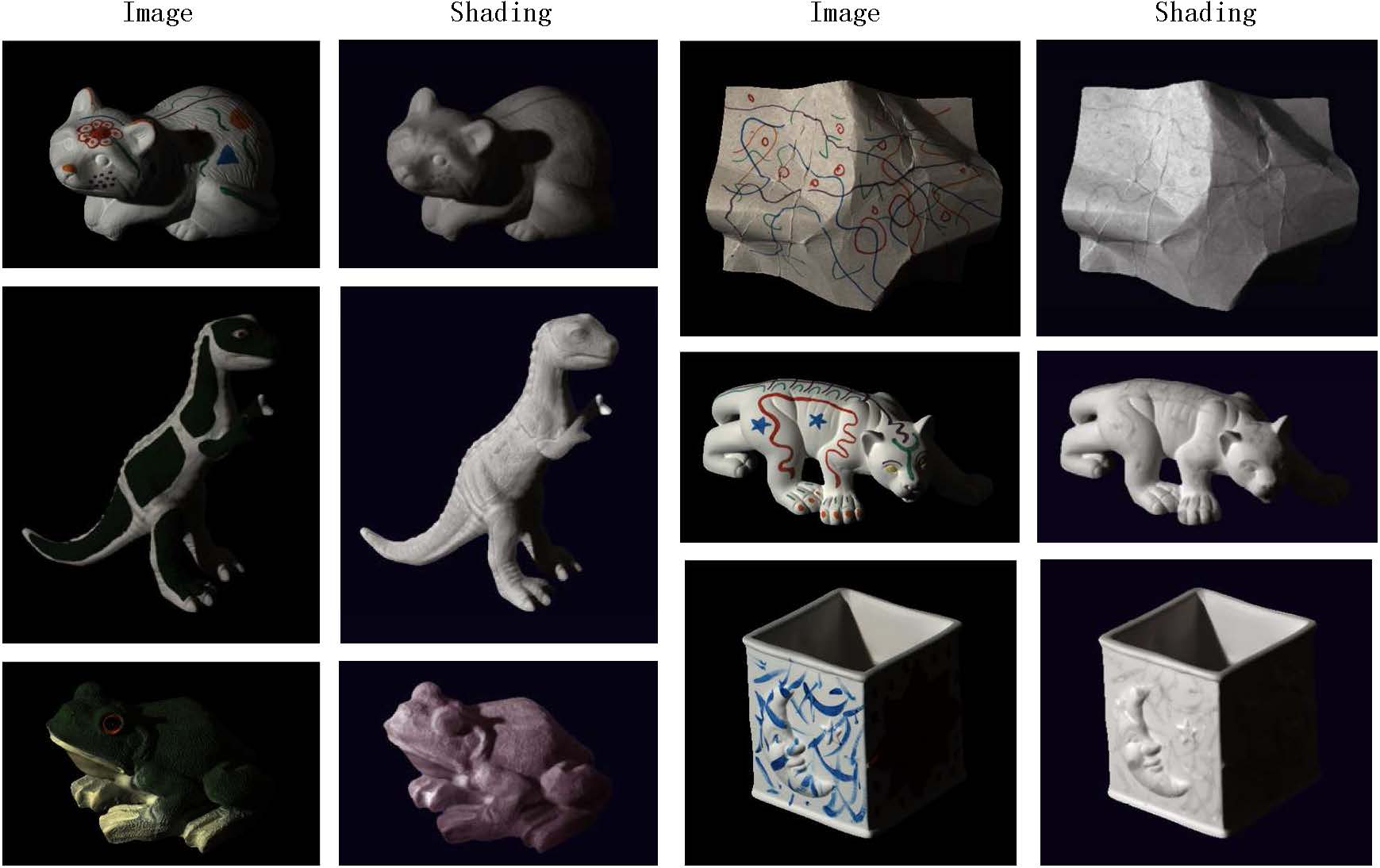}\\
  \caption{Shading colors in example images of the MIT intrinsic images dataset.
  }\label{fig:shading_color_MIT}
\end{figure}

Figure \ref{fig:shading_color_MIT} shows the colours of shading in images from the MIT Intrinsic Images dataset. We can see that most of the shading images are nearly achromatic. The reason is that the images are captured in controlled environment, where the ambient illuminations are largely suppressed by painting the background into black. According to Equation 3, when the ambient illumination is negligible, the shading will be nearly achromatic, no matter what the color of the direct illumination is. However, for the frog in Figure \ref{fig:shading_color_MIT}, the shading is slightly chromatic.

\begin{figure}
  \centering
  \includegraphics[width=1.0\linewidth]{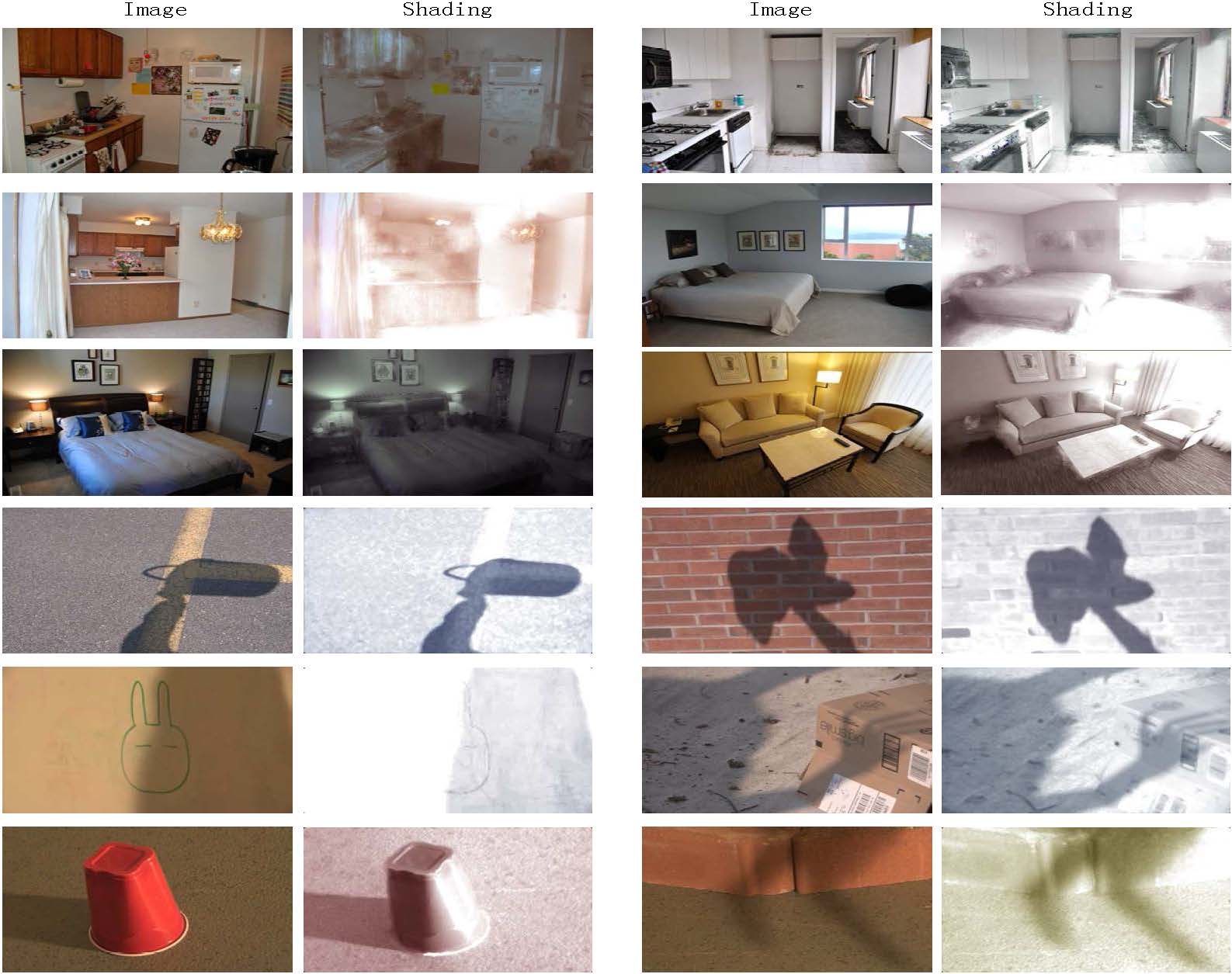}\\
  \caption{Shading colors in example images of natural indoor or outdoor scenes.
  }\label{fig:shading_color_indoor_outdoor}
\end{figure}

Figure \ref{fig:shading_color_indoor_outdoor} shows the colours of shading in natural indoor and outdoor scenes. The indoor scenes often have complex illuminations, so the shading colors vary a lot from image to image, even from place to place in the same image. In comparison, the shading colors in outdoor scenes are more regular. Especially, the shadows in outdoor scenes are often blueish, since the ambient light is often the blue sky.

%

%

%
%

\ifCLASSOPTIONcaptionsoff
  \newpage
\fi



\bibliographystyle{IEEEtran}
\bibliography{egbib}
%
%
%

%
\begin{IEEEbiography}[{\includegraphics[width=1in,height=1.25in,clip,keepaspectratio]{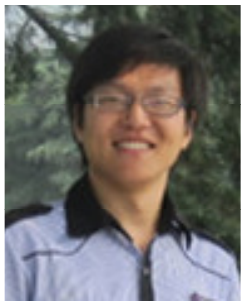}}]{Yuanliu Liu}
received his bachelor's degree and Ph.D. degree from Xi'an Jiaotong University, China, in 2008 and 2016 respectively. He was a visiting scholar in the University of Arizona in 2012-2013. His research interests focus on image processing and computer vision, especially on low-level vision.
\end{IEEEbiography}
\begin{IEEEbiography}[{\includegraphics[width=1in,height=1.25in,clip,keepaspectratio]{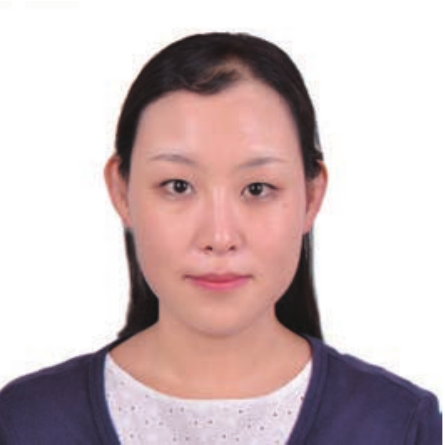}}]{Ang Li}
received the B.S. degree in  automation science from Xi'an Jiaotong University, China, in 2013. She is currently pursuing the Ph.D degree in  control theory and control engineering from the same university. Her research interests focus on image processing and computer vision.
\end{IEEEbiography}
\begin{IEEEbiography}[{\includegraphics[width=1in,height=1.25in,clip,keepaspectratio]{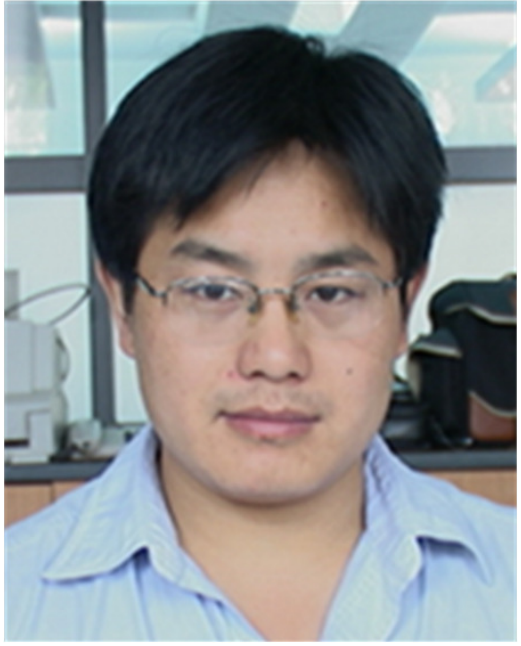}}]{Zejian Yuan}
received the MS degree in electronic engineering from the Xi'an University of Technology, Xi'an, China in 1999 and the PhD degree (2003) in pattern recognition and intelligent system from Xi'an Jiaotong University, China. Dr. Yuan was a visiting scholar in the Advanced Robotics Lab of Chinese University of Hong Kong in 2008-2009. He is currently an associate professor at the Department of Automatic Engineering, Xi'an Jiaotong University, and a member of Chinese Association of Robotics. His research interests include image processing, pattern recognition, as well as machine learning methods in computer vision.
\end{IEEEbiography}
\begin{IEEEbiography}[{\includegraphics[width=1in,height=1.25in,clip,keepaspectratio]{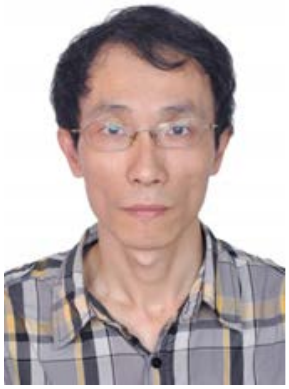}}]{Badong Chen}
received the B.S. and M.S. degrees in control theory and engineering from Chongqing University, in 1997 and 2003, respectively, and the Ph.D. degree in computer science and technology from Tsinghua University in 2008. He is currently a professor at the Institute of Artificial Intelligence and Robotics, Xi¡¯an Jiaotong University. His research interests are in signal processing, information theory, machine learning, and their applications in cognitive science and engineering. He has published 2 books, 3 chapters, and over 100 papers in various journals and conference proceedings. Dr. Chen is an associate editor of IEEE Transactions on Neural Networks and Learning Systems (TNNLS) and Journal of The Franklin Institute.
\end{IEEEbiography}
\begin{IEEEbiography}[{\includegraphics[width=1in,height=1.25in,clip,keepaspectratio]{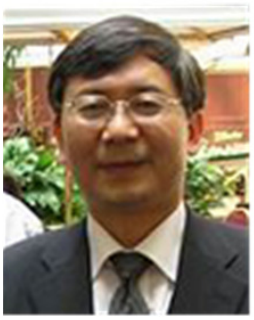}}]{Nanning Zheng}
graduated from the Department of Electrical Engineering, Xi'an Jiaotong University, Xi'an, China, in 1975, and received the M.S. degree in information and control engineering from Xi'an Jiaotong University in 1981 and the Ph.D. degree in electrical engineering from Keio University, Yokohama, Japan, in 1985.
He is currently a Professor and the Director of the Institute of Artificial Intelligence and Robotics, Xi'an Jiaotong University. His research interests include computer vision, pattern recognition and image processing, and hardware implementation of intelligent systems.
Dr. Zheng became a member of the Chinese Academy of Engineering in 1999, and he is the Chinese Representative on the Governing Board of the International Association for Pattern Recognition.
\end{IEEEbiography}

%
%
%




\end{document}